\pdfoutput=1
\documentclass[11pt]{article}

\usepackage[final]{acl}

\usepackage{times}
\usepackage{latexsym}

\usepackage{tcolorbox}
\usepackage{amsmath}
\usepackage{amssymb}
\usepackage{graphicx}
\usepackage{subcaption} % Use this for side-by-side figures

\usepackage{xcolor}
\usepackage{mdframed}
\usepackage{lipsum} % Just for sample text

\newmdenv[
  topline=True,
  bottomline=True,
  skipabove=\topsep,
  skipbelow=\topsep,
  leftmargin=0cm,
  rightmargin=0cm,
  innertopmargin=10pt,
  innerbottommargin=10pt,
  innerleftmargin=0pt,
  innerrightmargin=3pt,
  linewidth=2pt,
  linecolor=blue,
  backgroundcolor=gray!20
]{mybox}

\usepackage[T1]{fontenc}
\usepackage[utf8]{inputenc}
\usepackage{amsmath}

\usepackage{microtype}

\usepackage{inconsolata}

\usepackage{graphicx}

\usepackage{booktabs}

\title{Can Large Language Models \textit{Unlock} Novel Scientific Research Ideas?}

\author{Sandeep Kumar$\dagger$, Tirthankar Ghosal$\ddagger$, Vinayak Goyal$\dagger$, Asif Ekbal$\dagger$ \\
  $\dagger$Department of Computer Science and Engineering, Indian Institute of Technology Patna, India \\
  $\ddagger$National Center for Computational Sciences, Oak Ridge National Laboratory, USA\\
  $\dagger$\texttt{(sandeep\_2121cs29,2201ai52\_vinayak,asif)@iitp.ac.in}\\
  $\ddagger$\texttt{ghosalt@ornl.gov}}

\begin{document}
\maketitle
% \renewcommand{\thefootnote}{\fnsymbol{footnote}}
% \footnotetext[1]{\textit{Equal contribution}.}
% \renewcommand{\thefootnote}{\arabic{footnote}}
\begin{abstract}

% \textit{``An idea is nothing more nor less than a new combination of old elements"} \cite{young2019technique}\footnote{In this paper the old elements are the extensive general knowledge of LLMs, as well as the information drawn from related papers, which is presented in the related work and introduction sections. (See FAQ last point in  the Appendix)}. 

The widespread adoption of Large Language Models (LLMs) and publicly available ChatGPT have marked a significant turning point in the integration of Artificial Intelligence (AI) into people's everyday lives. This study examines the ability of Large Language Models (LLMs) to generate future research ideas from scientific papers. Unlike tasks such as summarization or translation, idea generation lacks a clearly defined reference set or structure, making manual evaluation the default standard. However, human evaluation in this setting is extremely challenging — it requires substantial domain expertise, contextual understanding of the paper, and awareness of the current research landscape. This makes it time-consuming, costly, and fundamentally non-scalable, particularly as new LLMs are being released at a rapid pace. Currently, there is no automated evaluation metric specifically designed for this task. To address this gap, we propose two automated evaluation metrics: Idea Alignment Score (IAScore) and Idea Distinctness Index. We further conducted human evaluation to assess the novelty, relevance, and feasibility of the generated future research ideas. This investigation offers insights into the evolving role of LLMs in idea generation, highlighting both its capability and limitations. Our work contributes to the ongoing efforts in evaluating and utilizing language models for generating future research ideas. We make our datasets and codes publicly available\footnote{\url{https://github.com/sandeep82945/Future-Idea-Generation.git}}.

\end{abstract}

\begin{quote}
\small
``Innovation is seeing what everybody has seen and thinking what nobody has thought''  ---Dr. Albert Szent-Györgyi
\end{quote}

\section{Introduction}
% What is an Idea-------------------
An \textit{idea} can be defined as a thought or suggestion aimed at solving a problem or considering a possibility. This concept is central to fields ranging from philosophy to science and economics. According to \cite{plato2000plato}, ideas are archetypal forms that represent the most accurate reality. In the context of scientific research, \cite{Kuhn1963TheSO} in "The Structure of Scientific Revolutions"  describes an idea as a realization or hypothesis that can challenge and shift paradigms within a scientific community. Therefore, an idea can be understood as a cognitive construct that arises from the human mind's ability to process information, reflect, and imagine, serving as a cornerstone for creativity, problem-solving, and innovation. Idea generation can generally be understood as a state of focused internally-directed attention involving controlled semantic retrieval \cite{BENEDEK2014125}.
\begin{figure}[t]
\centering
`    \includegraphics[width=0.40\textwidth]{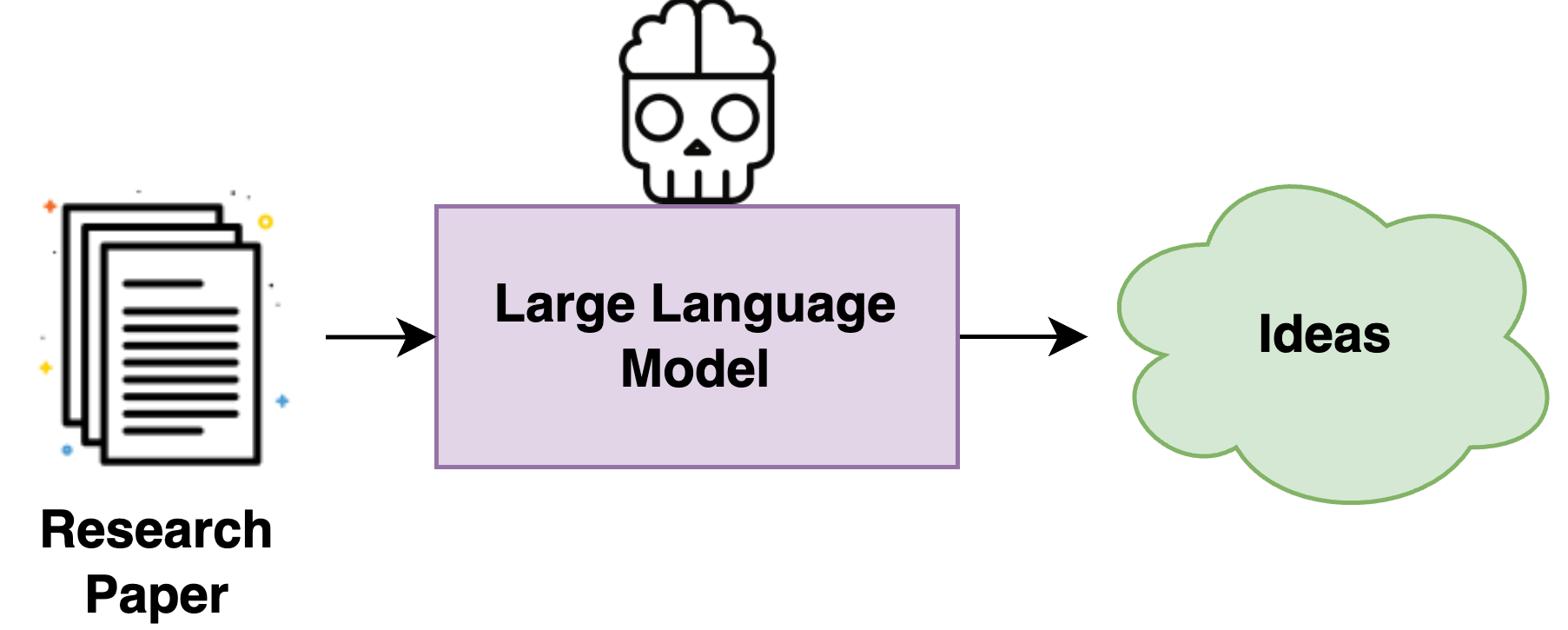}
    \caption{Large language model suggesting future research ideas after reading a research paper}
    \label{fig: sample}
\end{figure}

%----------Motivating towards large language can think like a human and it will be interesting to see if it can generate idea
%------ How large language can revolutionize idea generation
As technology improves, new capabilities emerge. Ever since the Turing Test was proposed in the 1950s, humans have explored the mastering of language intelligence by machine \cite{Zhao2023ASO}. Technological advancements serve two key functions in innovation. Firstly, they influence the goals of generating and selecting ideas. Secondly, they impact the methodology of how ideas are generated and chosen \cite{Kornish2017ResearchOI}. Large Language Models (LLMs) have exhibited unparalleled mastery of natural language processing (NLP). Since, these have become increasingly powerful, researchers have begun to investigate their reasoning ability in problem-solving tasks \cite{yao2022react,brahman2023making}. The concept of an idea is essentially a new combination of old elements. LLMs have access to a broad spectrum of knowledge, due to their extensive training on vast amount of text data. However, understanding how information extracted from a research paper can give rise to new ideas, which have not yet been explored much. This leads us to ponder:

\begin{tcolorbox}[colback=gray!10!white, colframe=gray!60!black]
Can Large Language Models read a scientific paper and suggest new research ideas or directions?
\end{tcolorbox}
%-----  Why we used future work to evaluate the research ideas

Motivated by this, in this paper, we analyze the potential of LLMs in generating future research directions or ideas. As LLMs possess knowledge across various domains, we investigate five specific areas, \textit{viz.} Computer Science, Physics, Chemistry, Economics, and Medicine. To address this task, we create a dataset of papers published post 2022 from these five domains. To the best of our knowledge, there is no existing automated evaluation metric specifically designed for assessing the quality of future research ideas (FRIs) generated by LLMs. Unlike tasks such as summarization or translation, idea generation lacks a clearly defined reference set or structure, making manual evaluation the default standard. However, human evaluation in this context is particularly challenging—it demands substantial domain expertise, a deep understanding of the source paper, and awareness of the current research landscape. As a result, it becomes time-consuming, costly, and fundamentally non-scalable, especially given the rapid pace at which new LLMs are being released. To evaluate the novelty and relevance of ideas generated by LLMs, we propose an Idea Alignment Score (IAScore). This score reflects how well the generated ideas align with those proposed by the authors. Our proposed IAScore provides a scalable, automated, and interpretable metric that captures how well a model can generate author-aligned research directions. While IAScore does not capture all novel ideas, it serves as a lower-bound indicator of a model’s competence. We annotate research papers with future research ideas to create a benchmark for evaluation. To assess the models' ability to generate diverse ideas, we introduce the Idea Distinctness Index. We further conduct a human evaluation of 660 generated ideas in the field of computer science to assess their novelty, relevance, and feasibility. Additionally, we analyze and discuss the performance and limitations of four LLMs: Gemini \cite{DBLP:journals/corr/abs-2312-11805}, Claude-2 \cite{Anthropic2023}, GPT-3.5, and GPT-4 \cite{DBLP:journals/corr/abs-2303-08774}. Our findings demonstrate that LLMs have the potential to generate research ideas that are relevant, distinct, feasible, and novel—to some extent.

To summarize, our main contributions in this
paper are as follows:
\begin{itemize}
    \item We contribute to the ongoing exploration of LLMs' capabilities in generating future research ideas.
    \item To address the task, we create a novel dataset of recent papers of five domains (Computer science, Economics, Chemistry, Physics, Medical).
    \item To assess the quality of generated ideas from LLMs, we propose Idea Alignment Score and Idea Distinctness Index to evaluate the idea generation capability.
    \item We discuss the challenges associated with human evaluation and conduct an in-depth analysis on the generated ideas.
\end{itemize}
% Our contributions are \textit{three-fold}: 1) We introduce a novel idea of assesing the idea generation capability of existing LLMs utlizing future work. 2) To address the task, we create a novel labeled dataset of around . 3) 

We hope that this work serves as a foundation for future studies focused on accelerating scientific research by automatically generating research ideas.

\section{Related Work}

Recently, LLMs have shown tremendous abilities to perform tasks they were not explicitly trained for \cite{DBLP:conf/iclr/WeiBZGYLDDL22, DBLP:journals/corr/abs-2303-12712}. This includes common sense question answering, code generation, and cross-domain problem solving, enriching their utility across unforeseen domains \cite{DBLP:journals/corr/abs-2107-03374,DBLP:conf/icer/SarsaDH022}. Their capability extends to advanced scientific domains such as computer science, physics, medicine, and mathematics \cite{RomeraParedes2023MathematicalDF, Huang2023BenchmarkingLL}. Technology Semantic Network (TechNet) was proposed to stimulate idea generation in engineering design \cite{Sarica2021IdeaGW}. There have been a few works in the discovery of new proteins to accelerate scientific discovery. The prior work reported in \cite{DBLP:conf/kdd/SpanglerWBNDHRPCMSKLLPLDCL14} involves utilizing published studies to find new protein kinases that phosphorylate the tumor suppressor protein p53.

A hypothesis is a hunch, assumption, suspicion, assertion or an idea about a phenomenon,
relationship or situation, the reality or truth of which you do not know \cite{kumar1996research}. There have been some works on hypothesis generation. Initial studies on automated hypothesis generation begin by constructing a corpus of distinct concepts. Subsequently, they explore the relationships between these concepts using machine learning techniques, such as analyzing the similarities among vectors representing different words (or concepts) \cite{DBLP:journals/nature/TshitoyanDWDRKP19}, or applying link prediction methods over a graph (where concepts are nodes) \cite{Nadkarni2021ScientificLM}. Recently, \cite{Qi2023LargeLM} used LLMs and extensive pre-existing knowledge of various scientific fields for hypothesis generation. PaperRobot \cite{wang-etal-2019-paperrobot} predicts related entities for an input title and writes key elements of a new paper, including the abstract, conclusion, and future work, and predicts a new title.

\citet{xu-etal-2023-exploring} developed a framework that leverages the concept co-occurrence graphs and a masked language model to explore and verbalize academic ideas. Their method involves constructing evolving concept graphs across various disciplines and utilizing temporal link prediction to identify potential interdisciplinary connections. The framework also incorporates pre-trained language models to articulate these connections in a coherent academic context. SciMON \cite{wang2023learning} showed that LLMs can be guided by seed terms to generate specific ideas. 

% They applied a pre-trained sentence classifier to classify sentences from the title and abstract into categories of {Background, Method, Objective}. They considered sentences labeled as problems or motivations as background, and the remaining were treated as target output sentences. Additionally, a pre-trained entity extractor was used to extract salient seed terms.

However, previous works primarily focused on developing methods (linking and explaining entities, which may not sufficiently capture the complexity or explain how LLMs can solve real-world problems) for idea generation, whereas our work exhaustively focuses on evaluating the capability of LLMs in generating research ideas. Our goal is to assess the inherent ability of LLMs to generate future research ideas/directions.

\section{Dataset} \label{Sec: dataset}
Our dataset creation involves three steps: (1) Dataset Collection, (2) Future Research Ideas (FRI) Identification and removal, and (3) FRI generation.

\subsection{Dataset Collection}
We construct a corpus D by collecting 1,250 papers from S2ORC\footnote{250 papers from each domain.} from the domains of Computer Science, Economics, Physics, Chemistry, Medical from \cite{lo-etal-2020-s2orc}\footnote{They used Science Parse and Grobid tool to extract data from PDF. We used the plain text without figure and tables and references for our experiment.}. We prompted the LLM with paper text after removing the future work for generating future ideas.

To ensure the quality and relevance of the data and to utilize the future research ideas mentioned in a paper, the selected papers must meet the following requirements: (1) the paper must contain the full content, and (2) the paper must include a section on future work.

% As a result of this construction process, the dataset D comprises structured data in JsonL format, with each sample representing an individual paper.

\subsection{FRI Identification and Removal}
We first identify and remove any potential research ideas mentioned throughout the paper. By doing this, we ensure that LLMs have no prior access to these ideas, which could otherwise affect the objectivity of the analysis. 
% We discuss the dataset collection in more detail in Appendix \ref{Section: dataset_collection}.

\subsubsection{Annotation Process}
Inspired by \citet{DBLP:conf/jcdl/HaoLQWZ20}, we define a future research idea as a discussion that the authors believe they will conduct in the future or believe needs to be investigated in future research. We discuss more details about the annotation guidelines, annotation training, annotation process, and annotator's pay in Appendix \ref{Appendix: A}.

% For the detailed annotation guidelines, please refer to Appendix \ref{sec:guidelines}.

\subsubsection{Future Work Removal}

We observed two types of future research ideas (FRIs)\footnote{In this paper, we use the terms `ideas,' `research ideas,' `future research ideas,' and `FRI' interchangeably to frequently refer to future research ideas.} (Direct FRI and Mixed FRI). We discuss them in  details in Appendix \ref{Appendix: Types_FRI}.

\noindent \textbf{AP-FRI Corpus: }
We removed the sentence from the paper's input text if it pertains to Direct FRI. However, in the case of Mixed FRI, we did not entirely remove the sentences; instead, we eliminated only parts of sentences or markers indicating future research ideas. We add the removed future ideas to a corpus, which we refer to as the AP-FRI (Author Perspective Future Research Idea Corpus). This corpus contains the future research ideas proposed by the authors of the paper. Also, before adding to the AP-FRI corpus, we merged the sentences about the same topic into a single group. We illustrate this by an example in Figure \ref{fig: editing}, mentioned in the Appendix of the paper.

\subsection{Data Statistics}
\begin{figure}[!ht]
\centering
    % First subfigure
    \begin{subfigure}{0.23\textwidth}
        \centering
        \includegraphics[width=\textwidth]{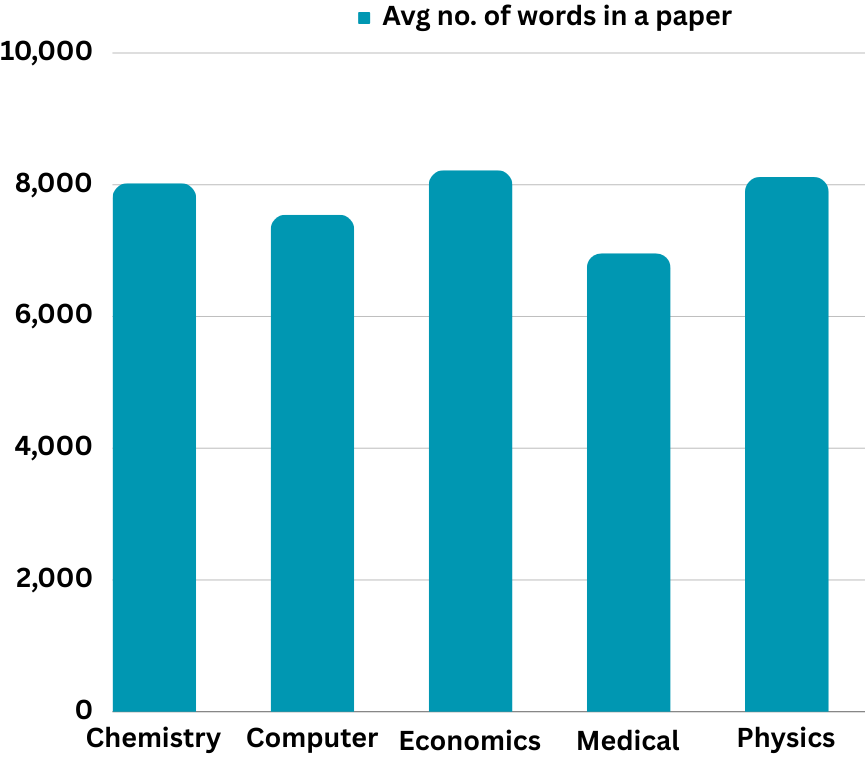}
        \caption{Domain vs Avg. number of words in a paper w/o FWK}
        \label{fig: stats_paper-a}
    \end{subfigure}
    \hfill
    % Second subfigure
    \begin{subfigure}{0.23\textwidth}
        \centering
        \includegraphics[width=\textwidth]{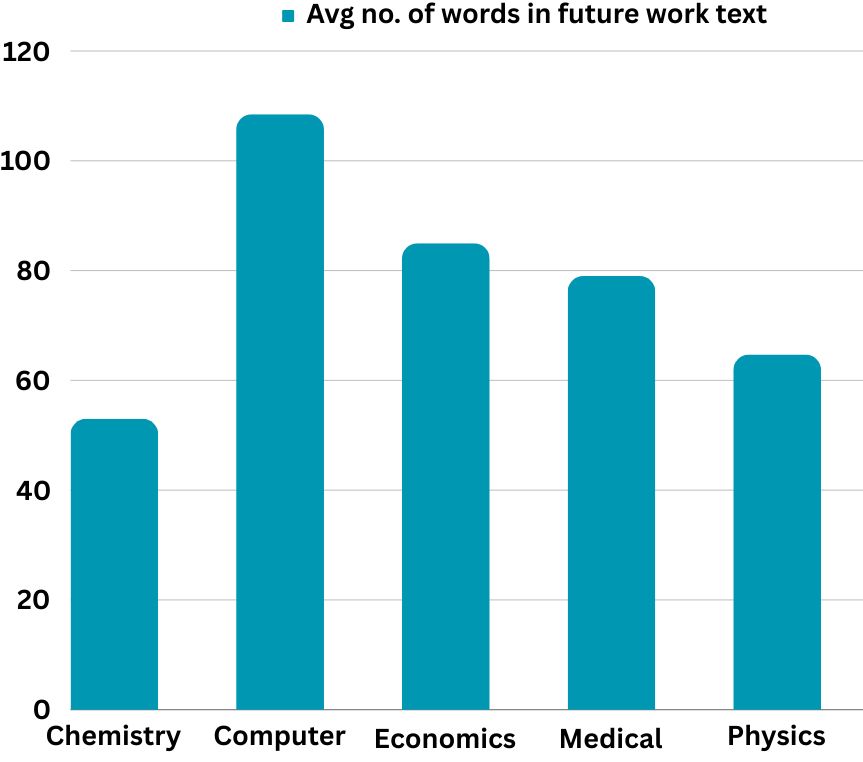}
        \caption{Domain vs Avg. number of words in FWK}
        \label{fig: stats_paper-b}
    \end{subfigure}

    \caption{Comparison of average word counts in papers with and without FWK across domains}
    
\end{figure}

% \begin{figure}[!ht]
% \centering
%     \includegraphics[width=0.30\textwidth]{full_text2.png}
%     \caption{Domain vs Avg. number of words in a paper w/o FWK}
%     \label{fig: stats_paper}

% \end{figure}

% \begin{figure}[!ht]
% \centering
%     \includegraphics[width=0.30\textwidth]{future_work1.png}
%     \caption{Domain vs Avg. number of words in FWK}
%     \label{fig: stats_future}

% \end{figure}

Figures \ref{fig: stats_paper-a} and \ref{fig: stats_paper-b} present a domain-wise analysis of average word counts in academic papers and their future work (FWK) sections, revealing significant variations across disciplines. Notably, Computer Science emphasizes future research extensively, while Chemistry adopts a more concise approach. Refer to the appendix Section \ref{Appendix: data_stats} for more details.

\section{Experiments}

\subsection{FRI Generation using LLM}
We investigate various prompts and utilize the following prompts to generate FRIs from the paper text after removing the FRIs mentioned in the paper as discussed in Section \ref{Sec: dataset}.
\begin{tcolorbox}[colback=gray!10!white, colframe=gray!60!black]
    \small
    \textbf{System}: You are a research scientist. \\
    \textbf{User}: Imagine you are a research scientist. 1) Read the full paper and understand it. 2) Find out the related works in this direction 3) Brainstorm and follow a step-by-step reasoning approach to generate potential future research ideas: \begin{quote}
        [paper text]
    \end{quote}
    Make sure the future research ideas are very distinct from the related papers. Potential future research ideas from the paper in bullet points are:
\end{tcolorbox}
Here, `[paper text]' contains the full content of the paper after removal of future work sections. In case the text was larger than the context size of the particular LLM we divided the paper into parts and finally combined the ideas.

\subsection{Challenges}
To accurately assess the novelty, relevance, and applicability of ideas generated by LLMs, evaluators must possess a high level of expertise in the specific domain and a deep understanding of the research topic to fully grasp the context. Additionally, they need knowledge of related literature to evaluate the ideas' future potential and the broader implications of their implementation.

\subsection{Idea Alignment Score (IAScore)}
With the above challenges, the evaluation of ideas generated by LLMs is a challenging process that demands a high number of domain-specific experts. We, therefore, proposed an \emph{Idea Alignment Score} (\( \text{IAScore} \)), which reflects how well the generated ideas align with those proposed by the author. The underlying idea for this score is that authors of accepted papers can be regarded as experts in their respective subjects. The reason being that they possess thorough background knowledge and have conducted deep analyses of the research topic before getting the paper accepted. Consequently, they are well-acquainted with the pertinent challenges which also may have been discussed by expert reviewers. Therefore, we propose that future ideas mentioned by the authors in the paper could be utilized as good quality of potential FRIs. 

% Utilizing these ideas, we define the Idea Alignment Score (IAScore) as follows:

The \text{IAScore} quantifies the alignment of newly generated ideas with author's perspectives within a specific domain, %we introduce a \emph{Idea Alignment Score} (\( \text{IAScore} \)). This score 
and is computed via a two-step process, detailed in Equations \ref{Eq: 1} and \ref{Eq: 2}.

Initially, we compute the average alignment score \( \text{AvgScore}_j \) for each paper's ideas. The \texttt{IdeaMatcher} model measures the alignment between the paper's author \emph{Future Research Ideas} (\( \text{AP-FRI}_j \)) and its each generated idea \( \text{I}_{ij} \). The subscript \( i \) indexes the \( i \)-th idea within the \( j \)-th paper, where \( N_j \) represents the total number of ideas proposed in that paper.

\begin{equation} \label{Eq: 1}
\small
\text{AvgScore}_j = \frac{1}{N_j} \sum_{i=1}^{N_j} \texttt{IM}\left( \text{AP-FRI}_j, \text{I}_{ij} \right)
\end{equation}

% \begin{equation} \label{Eq: 1}
% \begin{split}
% \text{Matcher}_{ij} &= \texttt{IdeaMatcher}(\text{AP-FRI}_j, \text{I}_{ij}) \\
% \text{AvgScore}_j &= \frac{1}{N_j} \sum_{i=1}^{N_j} \text{Matcher}_{ij}
% \end{split}
% \end{equation}

LLMs may generate new ideas that even the author may not have thought of. They can also generate additional future ideas, which may or may not be useful. Our goal with this score is to ensure that the LLMs generate potential future research ideas originally proposed by the authors\footnote{We intentionally avoided constraining the LLMs by not specifying a fixed number of ideas, allowing them to freely generate additional ideas beyond those proposed by the authors.}. Therefore, in our formula of \(\text{AvgScore}_j\), the sum of the alignment scores for a paper's ideas is divided by the total number of the author's proposed ideas, \(N_j\), to normalize the score.

\begin{equation} \label{Eq: 2}
\small
\text{IAScore}_{\text{domain, M}} = \frac{1}{P} \sum_{j=1}^{P} \text{AvgScore}_j
\end{equation}

Subsequently, we aggregate the individual paper scores to calculate the domain-wise \( \text{IAScore} \). This aggregation, presented in Equation \ref{Eq: 2}, averages the \( \text{AvgScore}_j \) values across all \( P \) papers within the domain. Higher the value of \( \text{IAScore}_{\text{domain}} \) signifies the more alignment of the generated ideas with author's perspective of all papers generated by model \( M \).

Let's define:
\begin{itemize}
    \item $I_{\text{valid}} \supseteq I_{\text{auth}}$: the set of all valid FRIs for a paper, including all author-defined FRIs.
    \item The probability that a generated idea $I \in I_{\text{gen}}$ is useful can be written as:
    \[
    \begin{split}
        P(I \in I_{\text{valid}}) &= P(I \in I_{\text{auth}}) \\
        &\quad + P(I \in I_{\text{valid}} \setminus I_{\text{auth}}).
    \end{split}
    \]
\end{itemize}

Now assume we cannot directly measure $I_{\text{valid}}$. The only known subset is $I_{\text{auth}}$, written by experts. So:

\begin{itemize}
    \item IAScore serves as a \textit{lower bound estimate} for $P(I \in I_{\text{valid}})$
    \item A model that scores poorly on the IAScore is unlikely to generate $I \in I_{\text{auth}} \subset I_{\text{valid}}$
\end{itemize}

\noindent Hence, if a model can not even match the known subset $I_{\text{auth}}$, we should not expect it to robustly hit the full space $I_{\text{valid}}$. IAScore is not intended to penalize novelty but to ensure a baseline level of credibility and expert alignment. It acts as a precision-driven benchmark: if a model can not even regenerate well-grounded ideas by domain experts, we cannot trust it to propose higher-quality novel ones. Hence, IAScore is a necessary but not sufficient component of idea evaluation and should complement---not replace---novelty-based costly human evaluation.

\subsubsection{IdeaMatcher}
% \begin{figure}[t]
% \centering
% `    \includegraphics[width=0.25\textwidth]{emnlp2023-latex/Idea matcher.png}
%     \caption{Research idea matching. AP: author predicted future research idea; Generated FRI: generated future research idea.}
%     \label{fig: idea_matcher}

% \end{figure}

To select an effective IdeaMatcher, we create a small annotated corpus. Our dataset was divided using the 30:70 ratio for validation and test sets, respectively. Since our study involves comparing two ideas using a pre-trained model, we did not require a separate training set. We first manually searched for matching pairs of ideas from generated ideas and AP-FRI of the paper. After obtaining 61 matching pairs, we searched for non-matching pairs of ideas, which is straightforward as only one generated idea will match or would not match with another one from AP-FRI while others would not match, so we picked an equal number of non-matching pairs. Then, we experimented with the idea-matching task by considering it similar to the Natural Language Inference (NLI) task. In particular, we considered the generated FRIs to be hypotheses and their corresponding AP-FRIs of the paper to be premises. If the idea matches, the hypothesis should be entailed by the premise. In particular, we used a pre-trained RoBERTa MNLI model \cite{liu2019roberta} for this task. We found that this technique produces many false negative cases, resulting in an accuracy of 65.5\%. 

We also evaluated the idea-matching capability of BERTScore \cite{bert-score}, as it utilizes BERT embeddings for comparison. We discuss the details in Appendix \ref{Appendix: BERT}. We found that BERTScore performed better than the entailment technique, resulting in an accuracy of 75.4\%. We also tried GPT by prompting it with various questions and found that it resulted in 91.8\% accuracy when prompted with a specific question prompt below:-
% We used the below prompt for this task:

% \begin{figure}[!t]
% \centering
%     \includegraphics[width=0.5\textwidth]{emnlp2023-latex/human_physics.png}
%     \caption{Novelty human evaluation for Physics domain}
%     \label{fig: human_physics}

% \end{figure}

\begin{tcolorbox}[colback=gray!10!white, colframe=gray!60!black]
\small
    \textbf{Prompt:}
        Your task is to examine whether a particular idea is incorporated within a set of ideas and to what degree. \\
        \textbf{Collection of ideas:} $\{ \text{API-FRIs} \}$ \\
        \textbf{Single idea:} $\{ \text{A generated Idea} \}$ \\
        Is the single idea contained within the collection of ideas? \\
        If yes, quantify its degree of presence or relevance of the single idea in the collection of ideas on a scale from $0$ to $1$.
\end{tcolorbox}

\begin{figure*}[!t]
\centering
    \includegraphics[width=0.55\textwidth]{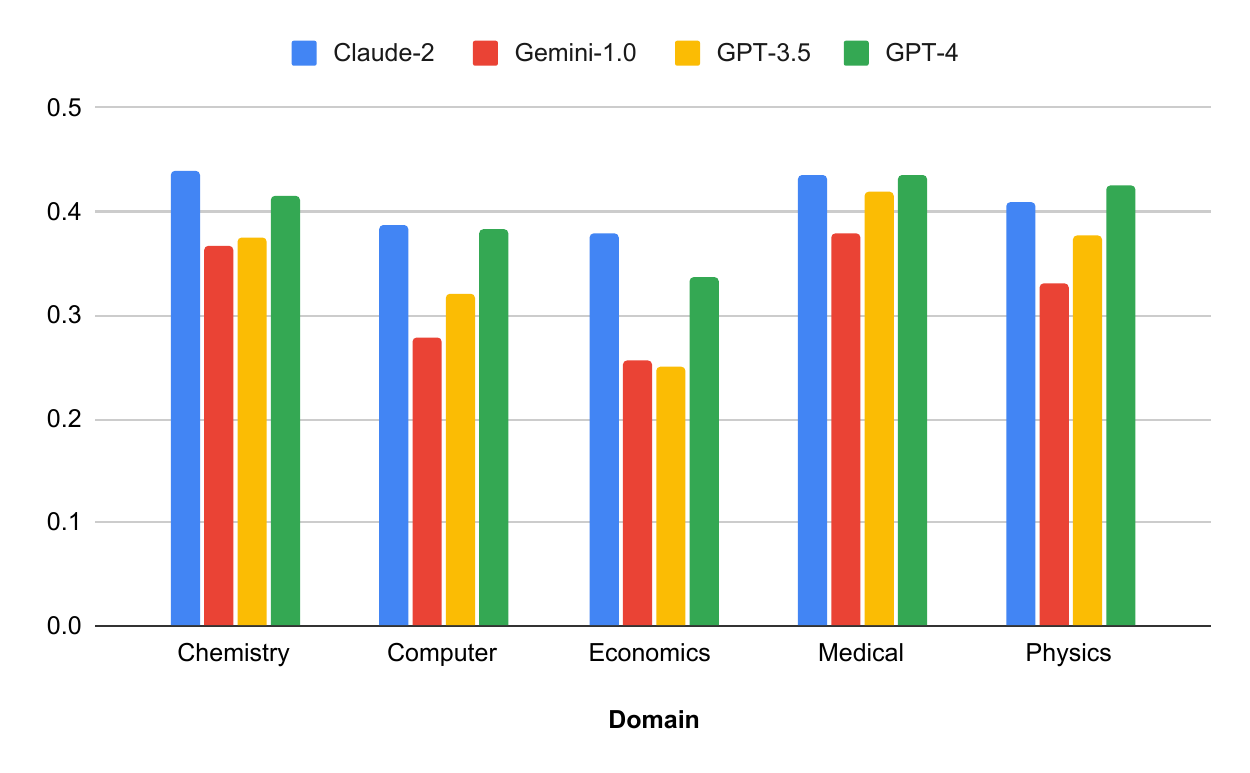}
    \caption{IAScore for each domain and model; a higher value indicates better alignment with the author.}
    \label{fig: friscore}
\end{figure*}

\begin{figure*}[!t]
\centering
    \includegraphics[width=0.6\textwidth]{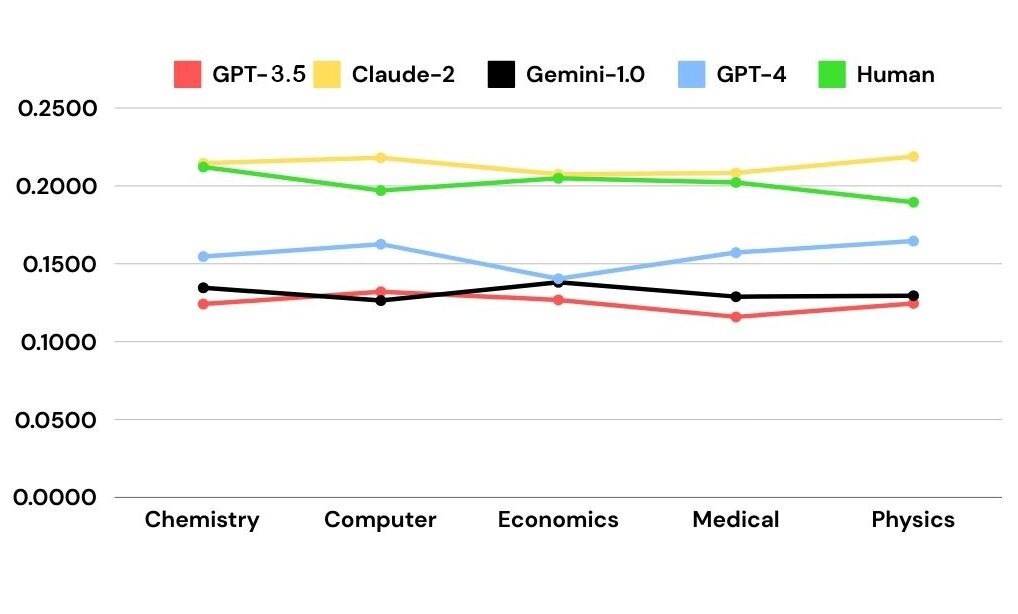}
    \caption{IdeaDistinctness index analysis; Here human is the authors of the paper}
    \label{fig: distinctness}
\end{figure*}

% \begin{figure}[!t]
% \centering
%     \includegraphics[width=0.45\textwidth]{emnlp2023-latex/human_final1.jpg}
%     \caption{Novelty human evaluation for Computer Science domain}
%     \label{fig: human_novelty}

% \end{figure}

\begin{figure}[!t]
\centering
    \includegraphics[width=0.50\textwidth]{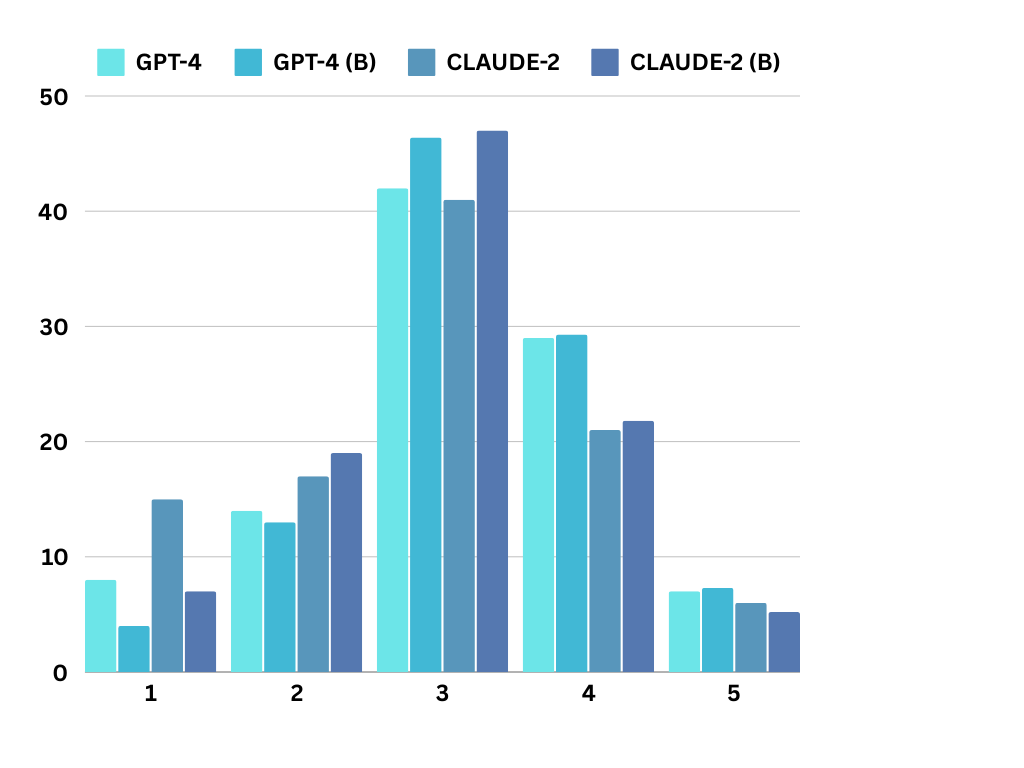}
    \caption{Novelty human evaluation for Computer Science domain. Here, (B) means with additional background knowledge; The x-axis represents the scale of novelty annotated by humans.}
    \label{fig: human_novelty}

\end{figure}

We found that GPT performs better than the existing NLI and similarity measure, such as BERTScore. Therefore, we chose GPT for this task\footnote{We used the OpenAI model GPT-3.5-turbo-0125 using OpenAI API. However, given the rapid advancements in language models, we recognize that newly released and upcoming models could further enhance the potential of IdeaMatcher.}

%as also shown in Figure \ref{fig: idea_matcher}.

\subsection{IdeaDistinctness Index}
Distinct-N \cite{Li2015ADO}, is a metric that measures the diversity of a sentence. It focuses on the number of distinct n-grams of a sentence, and thus penalizes sentences with a lot of repeated words. However, comparing two ideas need semantic comparisons rather than just syntactic differences. So, we introduce a method to semantically evaluate the distinctness of the generated ideas. This method in particular leverages semantic embedding to capture the essence of each idea and computes their distinctness based on semantic similarity measures. 

Given a set of generated ideas \( I = \{ \text{id}_1, \text{id}_2, \ldots, \text{id}_n \} \), representing individual ideas, we first encode each idea into a high-dimensional vector space using a pre-trained GPT embedding \cite{openai2022embedding}\footnote{text-embedding-ada-002} \(\mathbf{GPT}: id_i \mapsto \mathbf{v}_i\), where \(\mathbf{v}_i \in \mathbb{R}^d\) is the embedding of idea \(id_i\) and \(d\) is the dimensionality of the embedding space. 

To quantify the distinctness between pairs of ideas, we compute the cosine similarity between their embeddings, \(sim(\mathbf{v}_i, \mathbf{v}_j) = \frac{\mathbf{v}_i \cdot \mathbf{v}_j}{\|\mathbf{v}_i\| \|\mathbf{v}_j\|}\), for each pair of ideas \((id_i, id_j)\) in \(I\). The distinctness \(D_{ij}\) between two ideas \(i\) and \(j\) is then inversely related to their similarity: \(D_{ij} = 1 - sim(\mathbf{v}_i, \mathbf{v}_j)\).

The overall distinctness of the set \(I\) is calculated as the mean of all pairwise distinctness scores:
\begin{equation}
\small
D_I = \frac{1}{n(n-1)} \sum_{i=1}^{n} \sum_{j=1, j \neq i}^{n} D_{ij}
\end{equation}

This measure provides a single scalar value \(D_I\) that quantifies the average diverseness of ideas within a corpus of ideas, with higher values indicating a greater degree of diverseness among the ideas. 

Subsequently, we aggregated the distinctness scores across all ideas in each paper to compute the mean distinctness for that paper.
Let $P = \{p_1, p_2, \ldots, p_m\}$ represent the set of papers in a domain, where $m$ is the number of papers in the domain. Finally, for a comprehensive assessment of model performance within a domain, we averaged the mean distinctness scores of all papers generated by model \( M \) as follows:

\begin{equation}
\small
D_{\text{domain},M} = \frac{1}{m} \sum_{p=1}^{m} D_{I_{p_{M}}}
\end{equation}

The resultant metric, \( D_{\text{domain},M} \), represents the average idea distinctness for model \( M \) in a given domain, indicating the model's ability to generate diverse ideas.

We compute IA Score and Distinctness Index separately per domain to capture how LLM performance varies across disciplines. This design avoids conflating unrelated domains (e.g., chemistry vs. economics) and enables more granular, interpretable evaluation of LLM capabilities.

\subsection{Adding additional background knowledge}
\begin{figure}[ht]
\centering
    \includegraphics[width=0.49\textwidth]{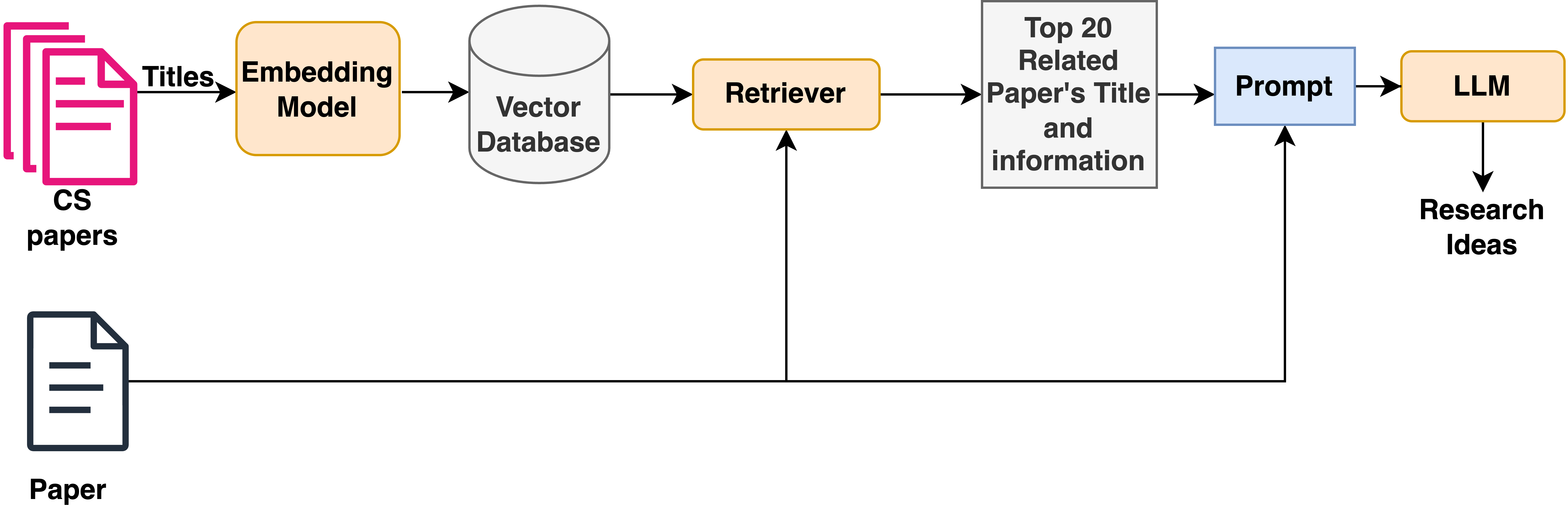}
    \caption{RAG Pipeline framework for infusing infusing more background knowledge with the LLMs}
    \label{fig: RAG}
     
\end{figure}

We designed our framework based on the Retrieval-Augmented Generation (RAG) model \cite{DBLP:conf/nips/LewisPPPKGKLYR020} to integrate background knowledge into LLMs, as illustrated in Figure \ref{fig: RAG}. A detailed explanation is available in the Appendix \ref{Appendix: background}, but we summarize the key components here.  1) \textbf{Vector Database}: We collected ~190K CS paper titles and abstracts via Semantic Scholar API \cite{Kinney2023TheSS} and stored BERT-based embeddings in a vector database. 2) \textbf{Retriever}: Computed cosine similarity with target title to retrieve top 20 related papers; used LLM prompts to extract their key contributions from abstracts. 3) \textbf{Generator}: Using the retrieved background knowledge, we prompted an LLM to generate distinct and novel future research ideas.

%We collected the titles and abstracts of around 1.9 lakh computer science research papers. Using BERT embeddings, we created vector representations of these titles and stored them in a vector database. From there, we retrieved the 20 papers most similar to our target paper's title. We extracted contributions from these papers' abstracts to gather relevant data and then generated ideas by prompting GPT-4 with the target paper and the retrieved background knowledge.

\subsection{Human Evaluation} \label{Sec: human}

The evaluation of generated future ideas necessitates familiarity with both previous works related to the subject and the work being evaluated. Specifically, the evaluator must be an expert in the domain and topic. Given the complexity of human evaluation, we approached authors (as the authors have the knowledge of their paper and they also have knowledge of the literature) who have published papers in reputable venues, possess over 5 years of experience in scientific publishing, and have authored more than 5 scientific papers. We collected their accepted papers (published between 2023 and 2024) and followed the process of dataset preparation as we discussed in Section \ref{Sec: dataset}, and generated FRIs. We modify the prompt slightly to specifically generate only the top five results (see Appendix \ref{Appendix: human}). We selected the outputs from Claude and GPT-4\footnote{We used gpt-4-turbo using OpenAI API for the generation} models due to their better IAScore and Idea Distinction index. We adopt this approach to avoid author exhaustion and to get an accurate evaluation. 
More details about the human evaluation are mentioned in the Appendix \ref{Appendix: human}.

\section{Results and Discussion}

% \subsection{Automatic Metrics}
% Rouge and Blue score
% However, similar to BLEU score, ROUGE score also has limitations. It may not fully capture the semantic meaning or coherence of the summary, and it relies solely on the n-gram overlap, which may not always be an accurate measure of summary quality.

\subsection{Alignment Results}
Figure \ref{fig: friscore} provides a comparative overview of the IAScore for four language models\footnote{We set maximum token length to 512, and temperature=0 for each model} Claude-2, Gemini-1.0, GPT-3.5, and GPT-4 across five academic domains: Chemistry, Computer Science, Economics, Medical, and Physics\footnote{We used GPT-3.5-turbo-0125 for GPT-3.5 (cutoff: December 2023), gpt-4-0125-preview for GPT-4 (cutoff: December 2023), anthropic.claude-v2 for Claude 2 (cutoff: August 2023), and gemini-1.0-pro-002 (cutoff: February 2023).}.

In the Chemistry and Economics domains, Claude has the highest IAScore, indicating strong alignment with the authors' future research ideas. Claude and GPT-4 have almost similar values for the Computer, Medical, and Physics domains (with GPT-4 slightly higher). GPT-3.5 and Gemini have lower scores than both GPT-4 and Claude in every domain. GPT-3.5 has almost the same score as Gemini in the Chemistry and Economics domains. However, it scores higher than Gemini in the Computer, Medical, and Physics domains. The results underscore the advancements in language model capabilities, with each model showcasing domain-specific strengths in idea generation. This alignment of LLMs shows that LLMs are able to generate relevant and novel ideas to some extent. We also studied the effect of length of future work on IAScore (See Appendix \ref{Appendix: length_effect}). Our overall analysis shows that ideas of moderate length (20-40 words) achieve the highest Impact Assessment Scores (IAScores), balancing detail and clarity, while shorter (<20 words) and longer (40-60 words) ideas tend to score lower due to insufficient detail. We also conducted a human analysis to understand the quality of research ideas generated when the IAScore is low (see Appendix \ref{Appendix: Error}).

\subsection{Distinctness Results} \label{Sec: idea_distinction}
We show the comparative evaluation of idea distinctness scores in Figure \ref{fig: distinctness}. The line graph depicts the variation of distinctness between the generated ideas and the human-written ideas (AP-FRIs). GPT-3.5 shows the least distinctness among the generated ideas, except in the Computer domain, where it is slightly more distinct than Gemini. As shown in the graph, the distinctness of Gemini is also quite low; however, it is slightly better than GPT-3.5, except in the Computer domain.

The generated ideas of GPT-4 are more distinct than those of Gemini and GPT-3.5 (except for economics, whereas the distinctness of GPT-4 is the same as Gemini). However, it is lower than both Claude and Human. The Idea Distinctness Index of the generated ideas from Claude are almost the same as those of humans for Chemistry, Economics, and Medical domains. However, they are higher than even human scores in the Computer and Physics domains, which shows that it generates very distinct FRIs.

% These calculations were repeated for each model across the respective domains of Chemistry, Computer Science, Economics, Medicine, and Physics. The comprehensive evaluation is illustrated in Figure X, which showcases the comparative analysis of idea diverseness for each language model across the studied domains.

% \usepackage{multirow}
% \usepackage{multirow}

% \usepackage{multirow}

\subsection{Human Evaluation Results}
We conducted a human evaluation on 660 generated ideas for 66 papers in the computer science domain. To validate the quality of human annotation, we measure the inter-annotator agreement ratio where 20\% of the generated ideas are evaluated by two different authors of the same paper. We measured Cohen’s kappa coefficient \cite{Cohen1960ACO}, which was 0.83, confirming the high-quality annotation of the research ideas generated. 

\textbf{Novelty:} Figure \ref{fig: human_novelty} displays the results of the human evaluation. We observed that Claude generates 14.78\% of non-novel and 16.52\% generic FRIs, 41.73\% moderately novel, 20.86\% very novel, and 16.52\% extremely novel FRIs. GPT generates 7.83\% not-novel, 13.91\% generic, 42.61\% moderately novel, 28.70\% very novel, and 6.96\% extremely novel ideas. Claude generates more non-novel and generic ideas than GPT-4, while GPT-4 produces more very novel ideas and nearly the same number of excellent ideas.

We conducted an additional human evaluation in the physics domain, analyzing 630 generated ideas across 63 papers. We observed that Claude generates 15.28\% non-novel, 16.02\% generic, 31.22\% moderately novel, 20.36\% very novel, and 17.12\% extremely novel FRIs. GPT produces 7.33\% not-novel, 14.53\% generic, 42.22\% moderately novel, 28.0\% very novel, and 7.95\% extremely novel ideas, similar to previous observations in computer science. This demonstrates that although LLMs also generate generic or already explored ideas, they are capable of producing novel ideas that have either not been explored or have been minimally explored\footnote{Comprehensive human evaluations across all domains require significant expertise and logistical resources, making it a challenge for the present study. However, the observed results strongly suggest that similar trends are likely to hold across other domains.}.

% Figure \ref{fig: human_novelty} displays the results of the human evaluation. We observed that Claude generates almost 14.78\% of not novel and 16.52\% generic FRIs, 41.73\% moderately novel, 20.86\% very novel and 16.52\% extremely novel. GPT generates 7.83\% of not novel, 13.91\% of generic, 42.61\% of moderately novel, 28.70\% of very novel and 6.96\% of extremetly novel ideas. Additionally, GPT-4 and GPT-3 generate almost 40\% of FRIs which already exist but are promising directions. 

\textbf{Relevance and Feasibility: } After human evaluation, for Computer Science domain, we found that 76.67\% of the ideas generated by Claude and 93.34\% by GPT-4 are relevant. Furthermore, 83.34\% of Claude's generated ideas and 96.64\% of GPT-4's ideas were judged to be practically feasible and factually correct. Similarly, for the Physics domain, after human evaluation, we found that 78.45\% of the ideas generated by Claude and 91.56\% by GPT-4 are relevant. Furthermore, 85.67\% of Claude's generated ideas and 94.78\% of GPT-4's ideas were judged to be practically feasible and factually correct.

These results highlight that Claude and GPT-4 can generate relevant and feasible research ideas. However, the reason Claude generates more impractical and irrelevant research ideas may be that Claude attempts to generate more distinct research ideas than GPT-4, as we evaluated and discussed in Section \ref{Sec: idea_distinction}.

\subsection{Open-ended generation:} 
We tested whether LLMs could retain open-ended generation capabilities by providing only a title and abstract as input. Our findings showed that, overall, LLMs can still generate open-ended content due to their past knowledge. However, they may not produce many high-quality ideas, as they lack access to recent publications and methodological insights relevant to the current paper. We discuss this in details in the Appendix \ref{Appendix: Open}.

\subsection{Impact on Adding Background Knowledge}
We found that adding background knowledge reduced the generation of generic or non-novel ideas and improved relevance and factual accuracy. However, further research is needed to boost the novelty of generated ideas. We discuss this in details in the Appendix \ref{Appendix: background}.

\subsection{Analysis on New Models}
\begin{figure}[ht]
\centering
    \includegraphics[width=0.50\textwidth]{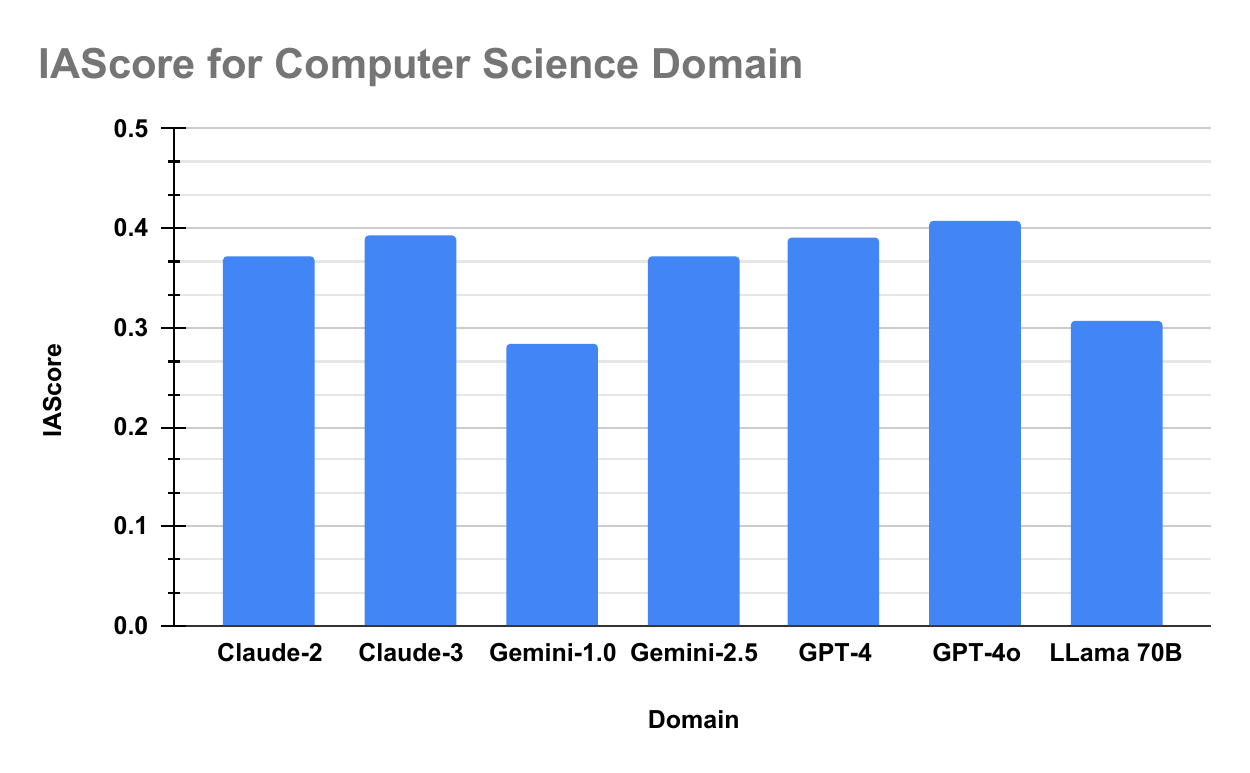}
    \caption{IAScore result for CS domain}
    \label{fig: iascore_new}
    
\end{figure}
As several new models were released after our initial experiments, we extended our analysis to include their results as well, as shown in Figure~\ref{fig: iascore_new}. GPT‐4o achieves the highest IAScore in the Computer Science domain (\(\approx 0.41\)), marginally surpassing GPT‐4 and Claude‐3 (both \(\approx 0.39\)). Claude‐2 and Gemini‐2.5 follow at \(\approx 0.37\), while Llama-70B and Gemini-1.0 trail at \(\approx 0.31\) and \(\approx 0.28\), respectively, illustrating the incremental gains delivered by successive model releases. We discuss more details on Appendix \ref{Sec: new_model}.

% \textcolor{blue}{We found the cases where GPT-4 can still generate few novel ideas by just providing title or abstract to the GPT-4 (see Appendix \ref{Appendix: Open} for the detailed discussion).}

% \subsection{LLM Evaluator}

% \subsubsection{Generic Ideas are getting generated}
% 1 paragraph

% \subsubsection{Is the Ideas factually correct?}

% \subsubsection{Is the generated ideas relevant?}

\section{Conclusion and Future Work}
In conclusion, we present the first attempt to evaluate the potential of LLMs in generating future research ideas across five domains: Computer Science, Economics, Chemistry, Physics, and Medicine. Here, we presented the first systematic attempt to evaluate the potential of LLMs for generating future research ideas across five diverse domains. By introducing the Idea Alignment Score (IAScore) and the Idea Distinctness Index, we offered scalable, interpretable metrics that complement costly human evaluation and provide a structured way to assess the quality of generated ideas. Our results and analysis show that LLMs possess domain-specific strengths in idea generation. Furthermore, the results from the Idea Distinctness Index indicate that LLMs, such as Claude and GPT-4, generate more distinct research ideas than Gemini and GPT 3.5. GPT-4 and Claude align better with authors written future research ideas than Gemini and GPT-4. The alignment of LLMs with the authors of generated ideas, and our human evaluation on relevance, novelty, and feasibility, reveal that although LLMs often produce non-novel and generic ideas, they have the potential to generate relevant and novel and diverse ideas to a significant extent. We hope that the findings and experiments of this work will unlock the potential of LLMs in idea generation and will foster new advancements in automated scientific innovation.

% We hope that the findings and experiments of this work will unlock the potential of LLMs in idea generation. and  will open a new topic of accelerating scientific research by automatically generating ideas.

In future, we plan to automate the extraction and annotation pipeline to enable scalable, automatic evaluation of our metrics for newly released models. Also, beyond relying on authors’ stated plans, future datasets could integrate ideas from related work, citing papers, and community discussions to reduce bias and capture broader research directions.

\section{Limitations} 
\subsection{Limitations of Data Collection} \label{Section: dataset_collection}
We extracted papers using the Semantic Scholar Academic Graph API from January 2023 to February 2024. The number of papers available is limited by the scope of our data extraction from the Semantic Scholar Academic Graph. We excluded papers that are not in English, as well as those whose abstracts could not be correctly parsed from the PDFs. Not all of these papers include sections on future work; therefore, we annotated only those that contained sections outlining future research directions. So due to such limitations, we collected 250 papers from each domain for analysis.

\subsection{Memorization} \label{Sec: Memoization}
\cite{Carlini2022QuantifyingMA} highlight that LLMs are prone to memorizing portions of their training data, a significant concern in the evaluation of contemporary LLMs. Despite this, the data used for pre-training and post-training includes "a small amount" of more recent data \cite{wang2023learning}. Therefore, we gathered recent papers from 2023 and 2024. By focusing our evaluation on papers published in these years,  the likelihood of test papers appearing in the pre-training corpora for the models is
substantially reduced. In addition, we conducted a manual review of these papers to assess memorization. This involved asking various questions related to the papers, such as their titles, publishing venues, author names, etc., to see if the models could supply the missing information. Our findings showed no evidence of such memorization occurring. A similar approach is also followed by \cite{wang2023learning} (discussed in Section 6.4) and even they did not find any evidence of this occurring.

\subsection{Limitation of IAScore and Idea Distinctness Index}
IAScore provides a systematic and interpretable method for measuring the alignment between ideas generated by LLMs and those identified by domain experts. While effective as a benchmark for assessing credibility and relevance, IAScore primarily reflects alignment with author-specified future research directions. Consequently, it may not fully capture innovative ideas proposed by LLMs that authors themselves have not explicitly mentioned. IAScore's performance is sensitivity to accuracy and comprehensiveness of the author-identified future directions. Moreover, IAScore emphasizes alignment rather than explicitly evaluating the depth, complexity, or practical feasibility of generated ideas.
 
As there is currently no reliable automatic metric for evaluating idea generation, human evaluation remains the primary approach. However, it is costly, resource-intensive, and not scalable. Our findings show that IAScore strongly aligns with expert judgments, but for more accurate and nuanced evaluation, human assessment is still preferred. Also, Idea Distinctness Index, which relies on cosine similarity, might miss subtle differences between ideas that appear semantically similar but actually represent distinct research directions.
\section*{Ethics Statement}
We have utilized the open source dataset for our work. Our aim for this work is to assess the potential of language models in generating ideas. Our Institutional Review Board (IRB) evaluated and approved this study. We do not encourage the use of LLMs to generate AI generated research papers (by generating new ideas) or misuse it for harmful idea generation. LLMs can process and synthesize vast amount of literature faster than humans, potentially identifying new patterns or gaps in research that might not be obvious, thus accelerating scientific discovery. However, since LLMs can generate content that may be similar to existing materials, this raises concerns about intellectual property rights and the originality of ideas. LLMs utilized for generating ideas might be misapplied to produce harmful materials such as plans for schemes for designs for destructive devices, explosive devices, ideas for spamming. Notably, it is a common challenge among existing LLMs with strong creative and reasoning abilities. So, we emphasize the responsible use of LLMs for idea generation and the need to broadly improve the safety of LLMs.

\section*{Acknowledgement}
Sandeep Kumar acknowledges the Prime Minister Research Fellowship (PMRF) program of the Govt of India for its support. We acknowledge Google for the "Gemma Academic Program GCP Credit Award", which provided Cloud credits to support this research. This research used the resources of the Oak Ridge
 Leadership Compute Facility (OLCF) at the Oak Ridge National Laboratory (ORNL), which is supported by the United States Department of Energy (DOE)’s Office of Science under
 Contract No. DE-AC05-00OR22725.

\bibliography{acl_latex}
\appendix

\section{Dataset Annotation} \label{Appendix: A}

\subsection{Dataset Annotation Guidelines} \label{Appendix: guidelines}
Recognizing future research idea in a paper involves analyzing the portion of text containing directions for future research. The following steps can be followed:

\textbf{Step 1:} Begin by reading the Title and Abstract of the paper to gain an understanding of its subject matter. It is important to read these sections multiple times to grasp the paper's main points, such as its motivation, contributions, and other relevant aspects. If necessary, refer to the paper itself or read related material to enhance your understanding.

\textbf{Step 2:} Identify Key Sections for Analysis
Focus primarily on the Discussion and Conclusion sections of the paper, as these areas often contain explicit mentions of future research directions. Scan the Methodology section as well, as sometimes suggestions for improving future studies or addressing current study limitations are mentioned here.

\textbf{Step 3:} Distinguish Future Research Ideas from General Statements:
Differentiate explicit future research suggestions from general discussion. Future research directions usually involve specific recommendations, plans, or identified gaps that require further exploration. These are often phrased using terms like "future studies should," "further research is needed," or "additional work will." Avoid confusing these with broader statements of potential relevance or applicability, which do not provide direct guidance on future work.

We offer multiple examples of papers with its future research ideas to assist and direct the annotators. We found a few text which looks like future work but is on contrary the motivation of the work. %For example in below example:-
As an example, consider the following:
\textit{"The goal of this work was to direct attention to emerging and novel research involving "magnetogel nanohybrid materials" that might be relevant in future applications for the treatment of wastewater, as well as in other fields.}

The second example is: \textit{"Our data could be useful for designing high-quality trials in the future to define the exact role of hemoadsorption in ARDS."}.

This sentence describes the future applications of existing research on magnetogel nanohybrid materials, not a specific direction for future research.

Also another example is: \textit{"The goal of this work was to direct attention to emerging and novel research involving magnetogel nanohybrid materials that might be relevant in future applications for the treatment of wastewater, as well as in other fields."} This is the application in future, and not the future work.

\textbf{Step 4:} Separate Future Research from Limitations:
Carefully examine any limitations mentioned in the paper to determine if they are explicitly linked to future research. Only consider a limitation as future work if the authors clearly indicate a direct intention to address it in subsequent studies. This helps avoid assuming that all limitations naturally lead to future research directions.

There is also very thin line between limitation and future work, where a limitation can or cannot be a future work. There were few cases where limitations were mentioned \textit{"One limitation of this paper is the absence of a coordinated attention structure to capture cross-channel information."}. As limitations can or cannot be a future work, we only take those limitations which is explicitly mentioned by the author to be a future work. Hence, we only considered the explicit mention of the future work by the author in their paper.

\subsection{Annotator Training}
Given the complexity of the papers and their frequent use of technical terminology, we hired two doctoral students, each boasting over four years of experience in scientific research publishing. To facilitate their training, an expert with more than ten years of experience in scientific publishing annotated 20 random papers from each domain, adhering to our guidelines. After this initial round of annotation, we reviewed and corrected any misinterpretations with the annotators, further refining their training and enhancing the clarity of our annotation guidelines. To assess the effectiveness of the initial training, we compiled another 20 papers from each domain. From the second round onward, the annotators consistently identified 95\% or more of the future research ideas correctly, based on comparison with expert-annotated labels.

\subsection{Annotation Process}
We regularly monitored the annotated data, placing emphasis on identifying and rectifying inconsistencies and cases of confusion. We also implemented an iterative feedback system that continuously aimed to refine and improve the annotation process. In cases of conflict or confusion, we removed those papers as we wanted only good quality dataset. Following the annotation phase, we obtained an average inter-annotator agreement score of 0.94 using Cohen's kappa \cite{Cohen1960ACO}, indicating a substantial consensus among the annotators.

\begin{figure}[t]
\centering
    \includegraphics[width=0.49\textwidth]{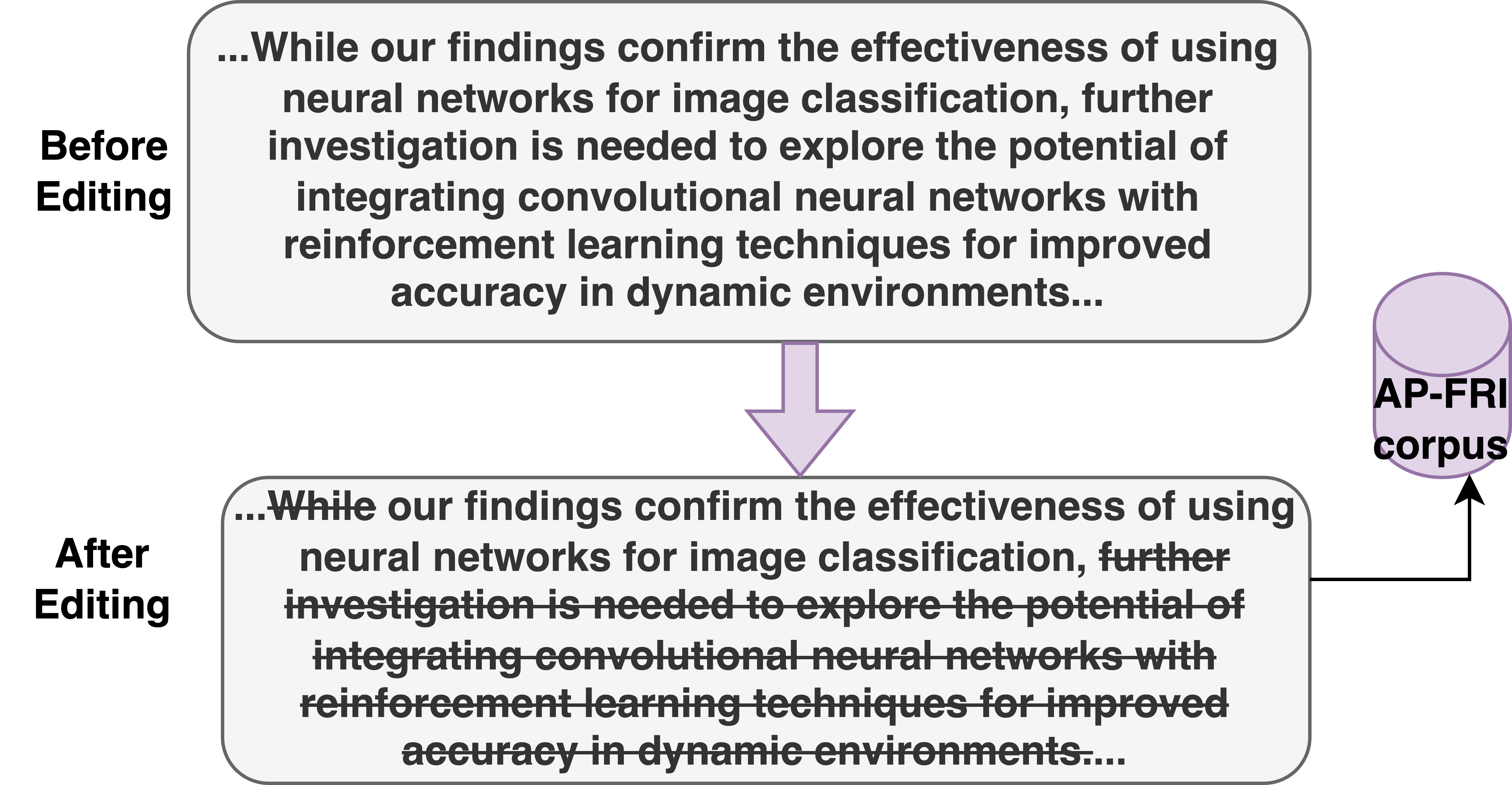}
    \caption{An example for FRI editing; Here the strike through text is removed from the paper text}
    \label{fig: editing}

\end{figure}

\subsection{Annotator's Pay}
We compensated each annotator according to the standard PhD salaries in India, based on the hours they worked. The appointment and salaries adhere to our university's established practices. Payment was made per paper since the time required to read and extract future research ideas from each paper varies, depending on its complexity, technical terminology, and the annotator’s familiarity with the subject. Thus, paying based on time spent could have potentially compromised the quality of the annotations. During the pilot phase, we observed that hourly-based payments encouraged speed over accuracy, potentially compromising annotation quality. In contrast, paying per paper ensured that annotators could dedicate the necessary effort and time required by each individual paper without worrying about time constraints or inefficiencies associated with clock-based metrics. In cases where papers were especially complex or ambiguous, we excluded them from the dataset rather than risk subpar annotation quality. To maintain accuracy and prevent fatigue, we imposed a daily limit of six hours for annotators.

\subsection{Detailed Dataset Statistics} \label{Appendix: data_stats}
Figure \ref{fig: stats_paper-a} provides a domain-wise distribution of the average word count in academic papers, excluding discussions on future work (FWK). It can be observed that the length of papers across all fields falls within a range of 7,000 to 8,000 words. Additionally, we calculated the average word count of extracted future work within each domain, providing comparative insights into how different fields prioritize discussions of future research directions. Figure \ref{fig: stats_paper-b} compares the average word count of future work text across six distinct scholarly domains. We observed that the literature in Computer Science notably prioritizes extensive discourse on future research, with an average word count significantly higher than that of other disciplines. In contrast, the literature in Chemistry demonstrates a more concise approach to discussions of future research, as evidenced by its lower average word count.

\section{Human Annotation}  \label{Appendix: human}

We ask the following questions from each human evaluator:-
\begin{itemize}
    \item Q1: Is the idea relevant with the research topic of the paper. (Relevant/Not relevant)
    \item Q2: Assess the originality/novelty of the research idea (5 scale)
    \item Q3: Review the research idea for factual correctness and feasibility. Is the idea impractical or too vague to be actionable? (Not Possible/Possible)
\end{itemize}
For Q2, we used Best-Worst Scaling \cite{louviere2015best} on a 5-point scale.

\begin{figure*}[ht]
\centering
    \includegraphics[width=0.7\textwidth]{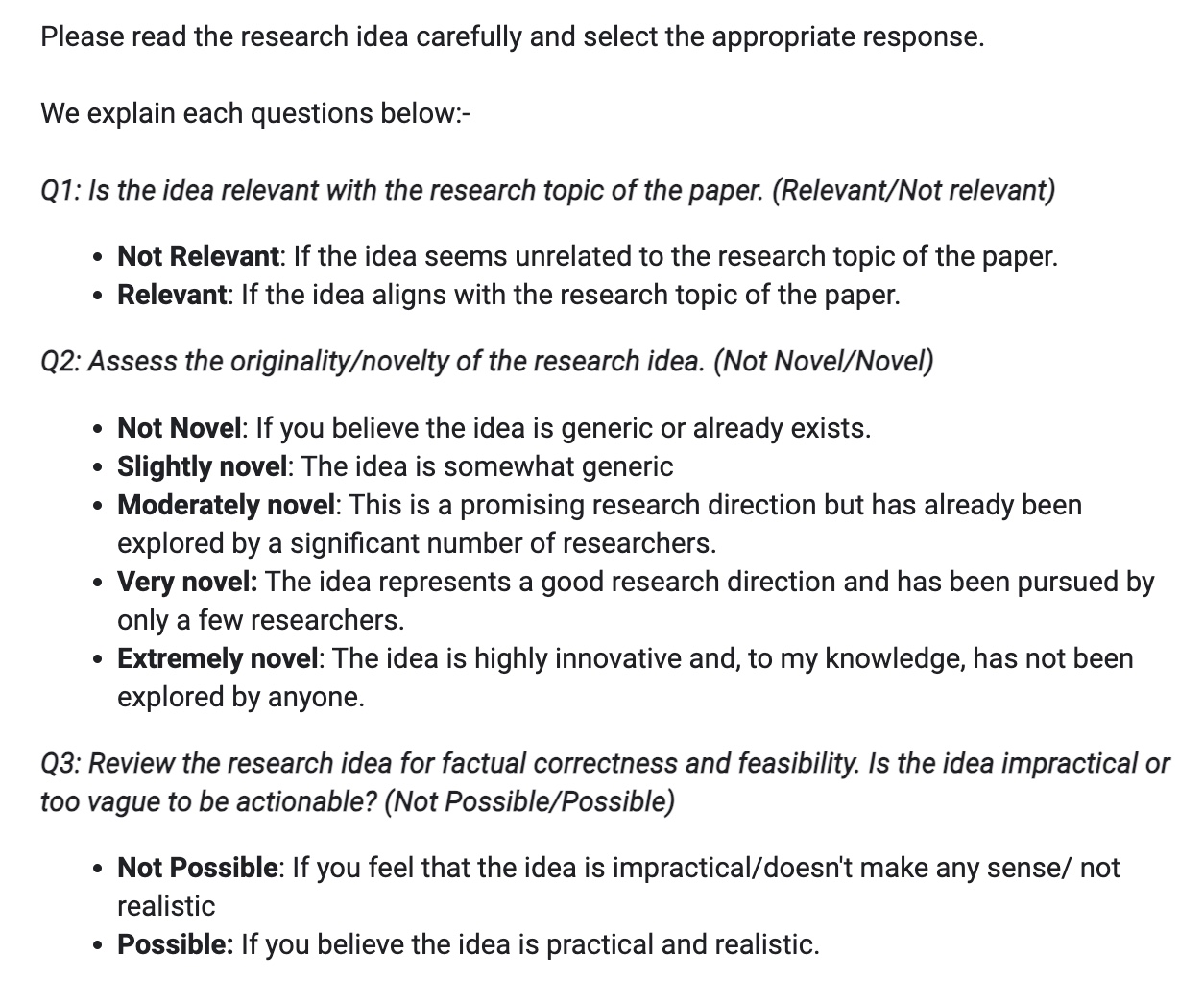}
    \caption{Instructions on Human Feedback form}
    \label{fig: feedback1}
\end{figure*}

\begin{figure*}[ht]
\centering
    \includegraphics[width=0.7\textwidth]{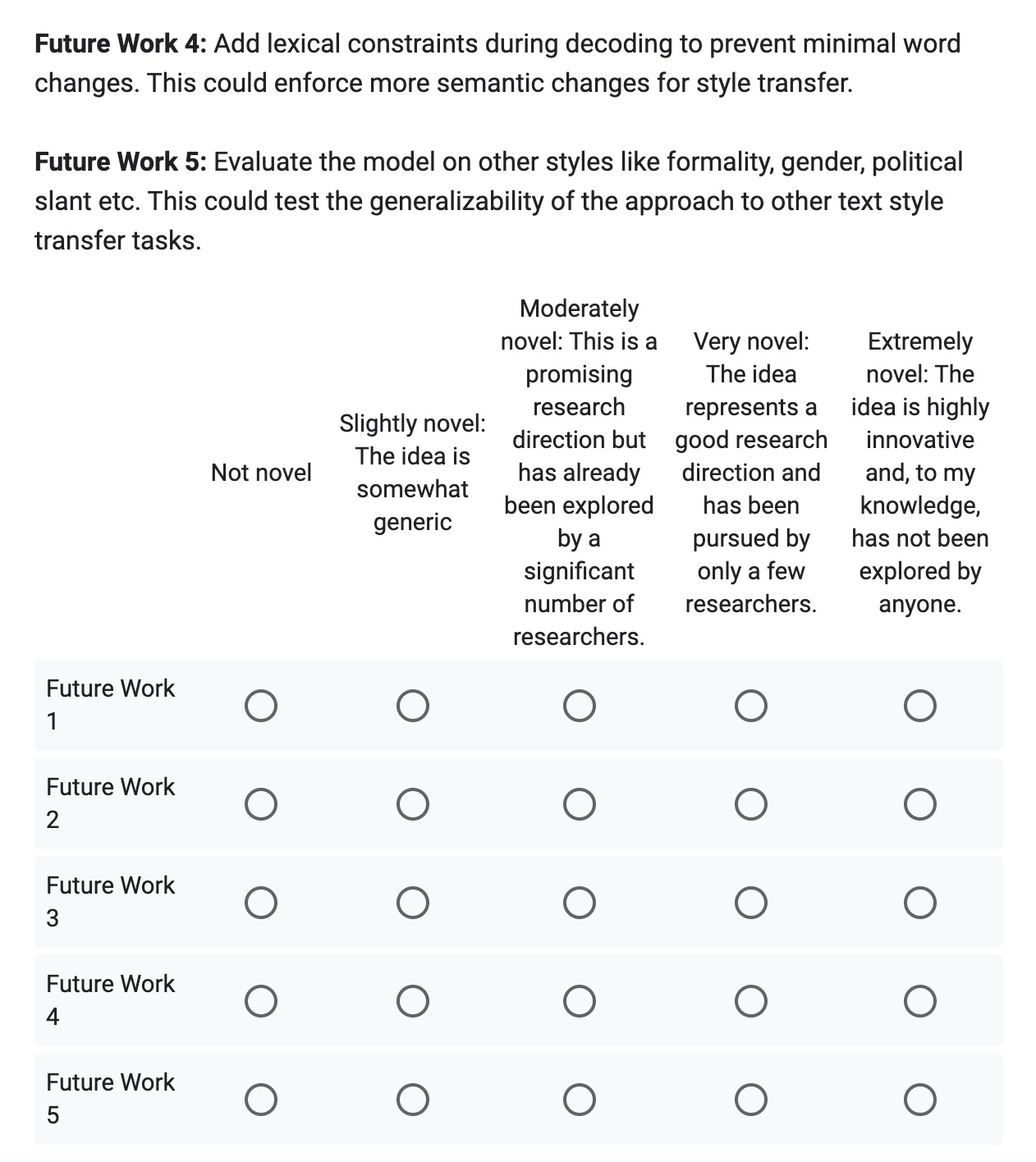}
    \caption{Human Feedback form for novelty}
    \label{fig: feedback2}
\end{figure*}

\begin{figure*}[!ht]
\centering
    \includegraphics[width=0.85\textwidth]{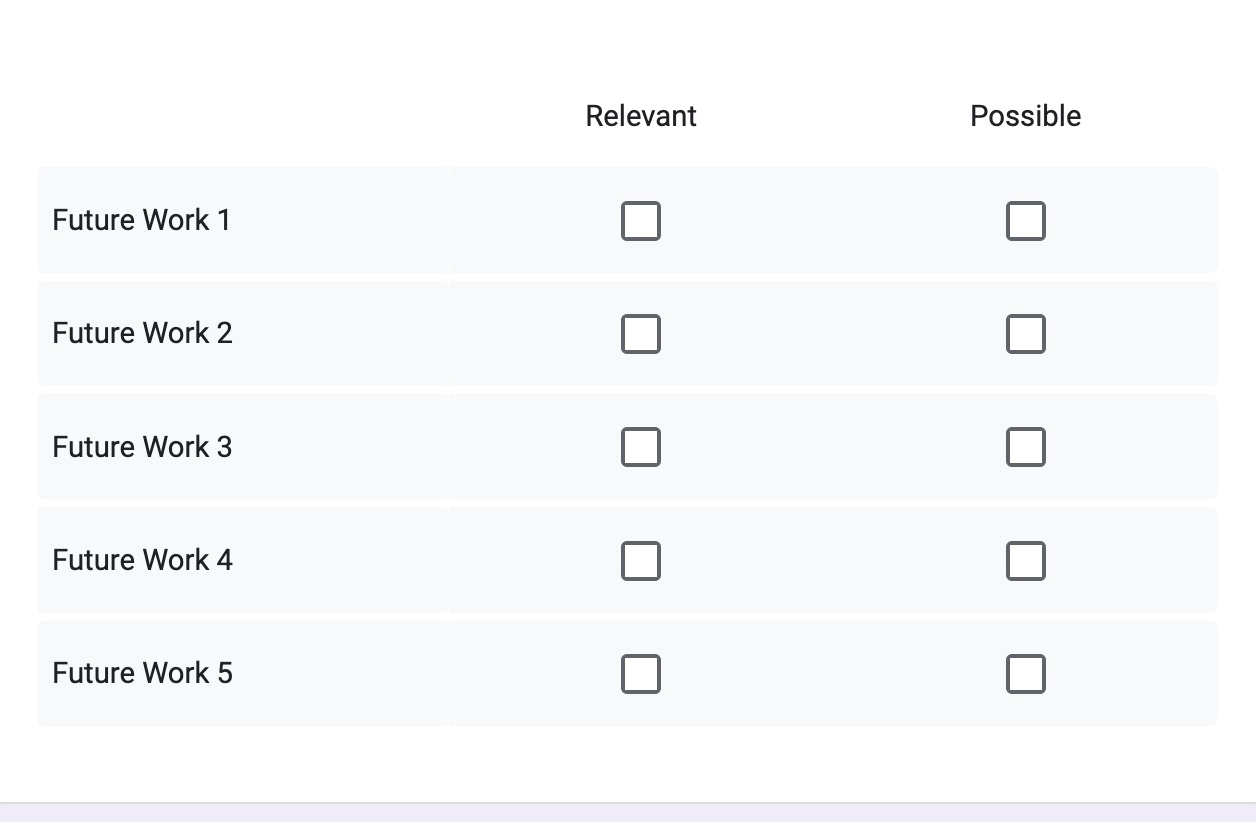}
    \caption{Human Feedback form for relevance, feasibility}
    \label{fig: feedback3}
\end{figure*}

We prepared a Google Form for each paper and provided the links to the annotators. We also specified instructions for them at the beginning of the form. We have added an example of the form for a paper in Figure \ref{fig: feedback1}, Figure \ref{fig: feedback2}, and Figure \ref{fig: feedback3}.

Here is the little modified from for human evaluation that generates only top 5 research ideas:-
\begin{tcolorbox}[colback=gray!10!white, colframe=gray!60!black]
    \textbf{System}: You are a research scientist. \\
    \textbf{User}: Imagine you are a research scientist. After reading the following paper, brainstorm to generate potential top 5 future research ideas:
    \begin{quote}
        [paper text]
    \end{quote}
    Potential top 5 future research ideas from the paper in bullet points are:
\end{tcolorbox}
Here, `[paper text]' contains the full content of the paper after removal of future work sections.

\section{Effect of giving only Title and Abstract as Input} \label{Appendix: Open}

We found a few cases where we provided only a title and abstract as input to see if LLMs can still retain open-ended generation capabilities. We discovered a few cases where GPT-4 still generated novel ideas, such as for a paper \cite{kumar-etal-2023-reviewers} it generated: \textit{"Incorporate explainable AI methods to provide transparency into how the AI model makes its predictions, thus making the outcomes more interpretable and acceptable to human editors."}. This kind of analysis has not been done yet and could be helpful. After providing full paper content to the model we found that same idea was again generated. 

There were also cases where GPT-4 generated a novel idea of solving the problem using transformers for a task (The task was mostly solved using techniques like RNN), which had not been done before. However, after providing the full paper content, the model understood that this transformer has already been implemented for this task, so further suggested to add more contextual information to it to boost the performance (limited information was given as input to the paper). Overall, we found that LLMs can still retain open-ended generation because it has past knowledge. But it may not generate many good ideas since it does not have access to recently published papers or other methodological findings related to the current paper.

\section{Effect of Length of Idea on IAScore} \label{Appendix: length_effect}
In our analysis, we explore the relationship between the length of ideas and their corresponding Impact Assessment Score (IAScore), specifically focusing on computer science papers and outputs generated by GPT-4. This relationship is visually represented in the bar chart found in Appendix Figure \ref{fig: lengthvsscore}. The data reveal that shorter ideas, typically under 20 words, tend to receive lower IAScores. This could be attributed to their lack of detailed information, which might be essential for a comprehensive understanding and assessment. Conversely, we observe that ideas spanning 40-60 words also tend to score lower. This may result from their verbosity; excessive information can dilute the core message, making it challenging to discern the main points. Interestingly, ideas with a moderate length, ranging from 20 to 40 words, achieve the highest IAScores. This length seems optimal as it allows for sufficient detail without overwhelming the reader, striking a balance that facilitates clearer understanding.

\begin{figure}[ht]
\centering
    \includegraphics[width=0.50\textwidth]{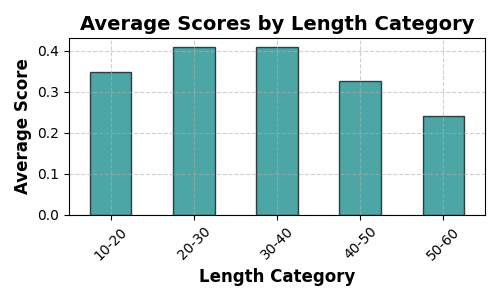}
    \caption{Effect of length on IAScore}
    \label{fig: lengthvsscore}
\end{figure}

\section{Effect of Adding Additional Background Knowledge}
\label{Appendix: background}

Below we describe our pipeline for retrieving background knowledge and generating future research ideas using LLMs in detail :-

\subsection{Vector Database}
We utilized the Semantic Scholar API \cite{Kinney2023TheSS} to collect the titles and abstracts of approximately 1.9 lakh existing computer science research papers. We employed BERT embeddings to create vector representations for the titles of these papers, which were then stored in a vector database.

\subsection{Retriever} To retrieve relevant papers, we created embeddings for the title of the paper for which we have to generate ideas. We computed the cosine similarity between the embedding of the title of this paper and those from our vector database. We then retrieved the top 20 research papers that exhibited the highest similarity to our target paper title. Finally, we extracted the contributions from these papers to gather relevant data from their abstracts.

We used the following prompt to instruct LLM to extract useful information from abstract of the paper:
   
\begin{tcolorbox}[colback=gray!10!white, colframe=gray!60!black]
    \textbf{System}: You are a helpful research agent that generates background knowledge or related works given abstracts of papers. \\
    \textbf{User}:  You are given abstracts of research papers and your task is to extract contributions or findings or methods proposed in the paper. You are not allowed to make any changes to data given to you. Return the response as it is and return response for all 20 papers in passage. Return title of paper followed by its contributions or findings or methods in less than 100 words. If no contributions or findings or methods are found, return NONE.
    \begin{quote}
        PASSAGE: '\{relevant\_passage\}'
    \end{quote}
    Potential top 5 future research ideas from the paper in bullet points are:
\end{tcolorbox}

We designed the above query prompt to ensure that the LLM\footnote{ We employed Gemini-Pro model for this task} understood its role in extracting relevant information without altering the provided information.

\subsection{Generator}
Next we produced the ideas using a prompt that includes the prompt using the paper and the retrieved background knowledge.

Specifically, we used the below prompt for our task:-

\begin{tcolorbox}[colback=gray!10!white, colframe=gray!60!black]
    \textbf{System}: You are a research scientist. \\
    \textbf{User}: Imagine you are a research scientist. 1) Read the full paper and understand it. 2) Find out the related works in this direction 3) Brainstorm and follow a step-by-step reasoning approach to generate  to potential future research ideas:
    \begin{quote}
        [paper text]
        [background knowledge]
    \end{quote}

    Make sure the future research ideas are very distinct from the background knowledge provided. Potential top 5 future research ideas from the paper in bullet points are:
\end{tcolorbox}
Here, `[paper text]' contains the full content of the paper after removal of future work sections. `[background knowledge]' contains the background knowledge retrieved. An example of background knowledge is shown in Appendix Table \ref{tab: background}.

We performed this experiment on the same set of papers and conducted human evaluations for novelty following the same methodology as we discussed in Section \ref{Sec: human}. The results are shown in Figure \ref{fig: human_novelty}. Initially, we observed that adding background knowledge affected the LLM's performance; it primarily generated ideas that already existed, merely creating new combinations from the background knowledge. Subsequently, we modified the prompt to instruct the model not to repeat ideas that were mentioned in the background knowledge. 

We found that adding background slightly improved the task. The results show that the improvements for GPT-4 and Claude were 50\% and 53.33\%, respectively, in reducing the generation of non-novel ideas. Also, it resulted in the improvement of 7.14\% and 11.76\% not generating generic ideas of GPT-4 and Claude. We observed that GPT-4 generated 9.52\% and 14.63\% more moderately novel ideas. However, we noted only a very slight improvement in the generation of highly novel or extremely novel ideas. 

The analysis revealed that 73.71\% of the ideas generated by Claude and 93.34\% by GPT-4 were relevant. We observed that the relevance score for Claude decreased by 2.96\%, and GPT-4 increased by a slight 0.77\%. Furthermore, 83.14\% of Claude's generated ideas and 96.98\% of GPT-4's ideas were judged to be practically feasible and factually correct. The score for Claude decreased by 0.20\%, and the score for GPT-4 increased by 0.34\%. It seems that additional information negatively impacts Claude's performance by generating ideas that are irrelevant, non-novel, and infeasible. However, for GPT-4, we observed that incorporating additional background knowledge helps prevent the generation of non-novel or generic ideas and slightly improves the relevance and factual correctness of the generated ideas. However, further research is needed to enhance the ability of LLMs to generate more novel ideas.

\section{BERTScore Implementation Details} \label{Appendix: BERT}

The motivation to use BERT embeddings is that the generated and the original ideas often do not use the same words, so we need to understand the contextual meanings of the ideas in order to compare them. We used the default setting of the BERTScore metric, which employs a 24-layer RoBERTa-large model and utilizes the 17th layer for embedding. We determined the threshold\footnote{We set the threshold 0.68 empirically} using the validation set. If the similarity exceeds that threshold, we classify those pairs of ideas as similar, and vice versa. 

\section{Error Analysis:} \label{Appendix: Error}

We conducted a human evaluation using three expert annotators, each with more than five years of experience in this field. They reviewed 15 papers. We assigned papers to each reviewer based on their familiarity with the subject matter of the papers. We identified two major reasons for the low IAS score:

\begin{itemize}
    \item Generic Ideas: Few ideas such as \textit{``Explore different explainability methods like LIME, SHAP to generate model explanations instead of just rationales. Compare their effectiveness."}, \textit{Building on the baseline model, future research could explore more advanced NLP models and techniques for contradiction detection.} are generated. These statements are true; however, they are very generic and are common.
    
    \item Not Feasible Ideas: We observed that LLMs occasionally propose ideas that are not feasible, even in principle. For instance, in one case, the model suggested ``\textit{collecting real-time, large-scale annotated data on private peer review discussions between authors and reviewers to study the evolution of scientific ideas.}” Such data is not only unavailable, but also extremely unlikely to ever be accessible due to strict confidentiality and privacy concerns in the peer review process. Human researchers, on the other hand, are typically more attuned to such ethical and practical constraints and tend to propose ideas that are realistically actionable.

    \item Author Miss: Due to page limits or more novel ideas, the author fails to mention a few ideas in a paper. For example, for a paper \cite{kumar-etal-2023-reviewers} GPT-4 generated idea: \textit{"Exploring the Impact of Contradictions on Review Outcomes: An interesting area for future research would be to study the impact of reviewer contradictions on the outcomes of the peer review process. This could involve analyzing the correlation between the presence and nature of contradictions and the final decisions made by editors (acceptance, rejection, major/minor revisions). Such studies could provide valuable insights into how contradictions influence the decision-making process and how they might be effectively managed to improve the fairness and quality of peer review."}. This represents a strong, novel research problem not mentioned by the authors, which warrants future investigation
\end{itemize}

\section{Direct FRI and Mixed FRI} \label{Appendix: Types_FRI}

\begin{itemize}
    \item \textbf{Direct FRI}: When the sentences that mention future research idea only contains future research idea. For example \textit{"In future work, we plan to extend our approach to other code-mixed languages and evaluate its performance on more NLP tasks."}
    
    \item \textbf{Mixed FRI}: We found that sometimes research papers articulate future research ideas along with other essential information of the paper in a single sentence. For example in Figure \ref{fig: editing}, this sentence not only summarizes the current research findings but also clearly outlines a direction for future work.

\end{itemize}

\begin{table*}[!ht] % Place the table on a separate page
    \centering
    \scriptsize
    \label{tab: example-1}
    \begin{tabular}{|p{1cm}|p{14cm}|}
        \hline
        \textbf{Model} & \textbf{Response} \\
        \hline
        Claude-2 & \begin{enumerate}
            \item Investigate the antioxidant and oil-stabilizing effects of peptides from the soy protein hydrolysates. The paper showed the pepsin hydrolysate had better activity, likely due to its peptide composition. Specific peptides could be isolated and tested for their antioxidant effects in oil.
            \item Study the effects of the soy protein hydrolysates on the quality and shelf life of the fried food. This paper focused only on the oil stability. It would be useful to also assess how the hydrolysates affect lipid oxidation, texture, color etc. in the fried banana chips.
            \item Evaluate the antioxidant effects of the soy protein hydrolysates in other vegetable oils such as sunflower, canola, corn oil etc. during frying. Since oil composition affects stability, the effects may differ in other oils high in polyunsaturated fatty acids.
            \item Examine the impacts of frying parameters such as temperature, food type, frying time etc. on the efficacy of the soy protein hydrolysates. Varying these factors could provide insights into the optimal conditions for the hydrolysates to exert antioxidant effects.
            \item Study the sensory qualities of fried foods with the soy protein hydrolysate additions. An analysis of taste, aroma, color acceptability by consumers would provide important information for potential commercial use of the hydrolysates as natural antioxidants.
        \end{enumerate} \\
        \hline
        Gemini-1.0 & \begin{enumerate}
            \item Investigate the effects of different hydrolysis conditions on the antioxidant activity of soy protein hydrolysates. This could involve varying the enzyme concentration, temperature, and pH of the hydrolysis reaction.
            \item Identify the specific peptides in the pepsin hydrolysate that are responsible for its antioxidant activity. This could be done using mass spectrometry and other analytical techniques.
            \item Test the antioxidant activity of soy protein hydrolysates in other food systems. This could include using them to stabilize other types of oils or to protect food products from oxidation.
            \item Investigate the potential health benefits of soy protein hydrolysates. This could include studying their effects on oxidative stress, inflammation, and other chronic diseases.
            \item Develop new food products that incorporate soy protein hydrolysates as antioxidants. This could include developing new types of oil-based products, such as salad dressings and mayonnaise, as well as new types of food products, such as fortified cereals and snacks.
        \end{enumerate} \\
        \hline
        GPT-4 & \begin{enumerate}
            \item Exploration of Peptide Sequences and Amino Acid Compositions for Enhanced Oxidative Stability: Investigate the specific peptide sequences and amino acid compositions derived from different proteases besides trypsin and pepsin to understand their contribution to the oxidative stability of oils. This could involve using a variety of proteolytic enzymes to hydrolyze soy proteins and other plant-based proteins, followed by an analysis of their peptide profiles and antioxidant activities in oil and food systems.
            \item Application of Soy Protein Hydrolysates in Various Food Systems: Extend the application of soy protein hydrolysates to other food systems beyond frying oils, such as baked goods, dairy products, and meat alternatives, to assess their potential in improving oxidative stability, shelf-life, and nutritional profiles. This could involve comparative studies of hydrolysate effectiveness in different food matrices and processing conditions.
            \item Mechanistic Studies on the Antioxidant Action of Hydrolysates: Conduct in-depth mechanistic studies to understand how soy protein hydrolysates interact with lipid molecules and radical species at the molecular level. This could involve computational modeling and in vitro experiments to explore the dynamics of peptide-lipid interactions and the role of specific amino acids in scavenging free radicals and metal ions.
            \item Development of Environmentally Friendly and Health-focused Antioxidants: Given the interest in replacing synthetic antioxidants with natural alternatives, further research could focus on optimizing the production, purification, and application of soy protein hydrolysates as natural antioxidants. This could include exploring sustainable extraction methods, evaluating the safety and efficacy of hydrolysates in food systems, and assessing consumer acceptance.
            \item Comparative Analysis of Hydrolysates from Different Oilseed Proteins: Since the study focused on soy proteins, future research could explore the antioxidative potential of hydrolysates derived from other oilseed proteins, such as sunflower, canola, or hemp. This would involve a comparative analysis of their antioxidative activities, peptide compositions, and effectiveness in enhancing the oxidative stability of different types of oils and fats used in food processing and storage.
        \end{enumerate} \\
        \hline
        GPT-3.5 & \begin{enumerate}
            \item Investigate the impact of different proteases on the antioxidant properties of soy protein hydrolysates in stabilizing oils during frying cycles.
            \item Study the influence of peptide size and amino acid composition on the oxidative stability of oils when treated with protein hydrolysates.
            \item Explore the potential of specific amino acids, such as tryptophan, histidine, and methionine, in enhancing the antioxidant activity of hydrolyzed soy proteins in frying oils.
            \item Examine the effects of pH adjustments during enzymatic hydrolysis on the antioxidant performance of soy protein hydrolysates.
            \item Evaluate the long-term stability and effectiveness of pepsin and trypsin hydrolyzed soy protein in preventing oil oxidation in different frying conditions over extended periods.
        \end{enumerate} \\
        \hline
    \end{tabular}
    \caption{Example of generated response for chemistry domain. Paper: \url{https://www.ncbi.nlm.nih.gov/pmc/articles/PMC10528889/pdf/foods-12-03498.pdf}}
    \label{Table: output-1}
\end{table*}

\begin{table*}[p] % Place the table on a separate page
    \centering
    \scriptsize
    \label{tab: example-2}
    \begin{tabular}{|p{1cm}|p{14cm}|}
        \hline
        \textbf{Model} & \textbf{Response} \\
        \hline
        Claude-2 & \begin{enumerate}
            \item Investigate other machine learning models like recurrent neural networks for multi-round transfer learning in lung cancer detection. The paper focused on convolutional neural networks, but RNNs may capture temporal patterns in medical imaging data.

            \item Explore different loss functions and regularization techniques to further minimize negative transfer. The paper proposed custom loss functions, but there is room to experiment with other formulations. 

            \item Evaluate the multi-round transfer learning approach on diverse medical imaging datasets beyond lung cancer, such as brain MRI scans. This could demonstrate the generalizability of the method.

            \item Develop more advanced generative models like CycleGAN to synthesize intermediate domains between source and target. This could create higher quality bridging distributions. 

            \item Perform multi-round transfer learning in a low-data regime by leveraging data augmentation and semi-supervised learning techniques. This could show the approach is viable when target data is extremely scarce.
        \end{enumerate} \\
        \hline
        Gemini-1.0 & \begin{enumerate}
            \item Investigate the impact of different prioritization algorithms on the performance of MTL-MGAN.The paper uses a hybrid approach that merges modified 2D dynamic warping (M2DW) and Silhouette coefficient to prioritize source domains. Future research could explore other prioritization algorithms, such as mutual information, transfer entropy, or graph-based methods, to determine if they can further improve the performance of MTL-MGAN.
            \item Develop new loss functions to minimize negative transfer.** The paper proposes loss functions that are designed to minimize negative transfer in the aspects of domains, instances, and features. Future research could investigate other loss functions that can further reduce negative transfer, such as adversarial loss, maximum mean discrepancy, or knowledge distillation.
            \item Explore different architectures for the MGAN.** The paper uses a modified GAN (MGAN) to create two intermediate domains as bridges between the source domain and the target domain. Future research could explore other GAN architectures, such as deep convolutional GAN (DCGAN), Wasserstein GAN (WGAN), or StyleGAN, to determine if they can further improve the performance of MTL-MGAN.
            \item Apply MTL-MGAN to other medical diagnosis tasks.** The paper evaluates the performance of MTL-MGAN on 10 benchmark datasets related to lung cancer detection. Future research could apply MTL-MGAN to other medical diagnosis tasks, such as breast cancer detection, skin cancer detection, or diabetic retinopathy detection, to determine its generalizability.
            \item Investigate the use of MTL-MGAN for other types of data.** The paper applies MTL-MGAN to medical images. Future research could investigate the use of MTL-MGAN for other types of data, such as text, audio, or time series data, to determine its applicability to a wider range of problems.
        \end{enumerate} \\
        \hline
        GPT-4 & \begin{enumerate}
            \item Extension to Other Forms of Cancer and Diseases: Investigating the applicability and effectiveness of the MTL-MGAN approach for the detection and diagnosis of other types of cancer, such as breast, prostate, or skin cancer, and potentially other diseases that have distinct imaging characteristics. This would involve adapting the transfer learning and MGAN components to handle different imaging modalities and disease signatures.

            \item Real-Time Diagnostic Systems Development: Developing real-time, AI-driven diagnostic systems using the MTL-MGAN framework that can be integrated into clinical workflows. This could significantly reduce diagnosis times and improve patient outcomes by providing immediate, accurate diagnostic information during patient examinations or surgeries.

            \item Integration with Electronic Nose Technology for Early Detection: Combining the MTL-MGAN approach with emerging technologies like electronic nose devices, which can detect volatile organic compounds associated with lung cancer from breath samples. Research could focus on creating a hybrid detection system that leverages imaging data and breath analysis for earlier detection of lung cancer.

            \item Exploration of Unsupervised and Semi-supervised Learning Techniques: Given the challenge of obtaining large, annotated medical imaging datasets, exploring unsupervised and semi-supervised learning methods within the MTL-MGAN framework could be valuable. This might involve developing algorithms that can learn from unlabelled data or very small labelled datasets, potentially increasing the model's applicability in low-resource settings.

            \item Improvement of Transfer Learning Techniques for Heterogeneous Data Sources: Investigating advanced transfer learning techniques that can effectively handle more significant variations between source and target domains, particularly when dealing with heterogeneous data sources. This research could include the development of novel algorithms for better domain adaptation and intermediate domain creation to facilitate the transfer of knowledge across diverse medical datasets and imaging modalities.
        \end{enumerate} \\
        \hline
        GPT-3.5 & \begin{enumerate}
            \item Exploring Multiround Transfer Learning in other Healthcare Applications: Investigate the applicability of multiround transfer learning in various healthcare domains beyond lung cancer detection, such as cardiovascular disease diagnosis, diabetic retinopathy detection, or infectious disease prediction.
            \item Advanced Loss Function Design for Negative Transfer Avoidance:Develop novel loss functions to further minimize the impact of negative transfer in transfer learning models by incorporating more sophisticated domain, instance, and feature-based penalties.
            \item Enhanced Generative Adversarial Networks for Bridging Domain Gaps:Investigate the development of advanced generative adversarial networks that can create more effective intermediate domains to bridge the gap between diverse source and target domains in transfer learning scenarios.
            \item Automated Prioritization Algorithm for Source Domains:Develop an automated algorithm that can dynamically prioritize source domains based on their similarities to the target domain, thereby enhancing the efficiency and accuracy of the transfer learning process.
            \item Evaluation of MTL-MGAN in Real-World Clinical Settings:Conduct a comprehensive evaluation of the proposed multiround transfer learning and modified generative adversarial network algorithm in real-world clinical settings to assess its performance, scalability, and generalizability in diverse healthcare environments.
        \end{enumerate} \\
        \hline
    \end{tabular}
    \caption{Example of generated response for computer domain. Paper:
    \url{https://www.hindawi.com/journals/ijis/2023/6376275/}}
    \label{Table: output-2}
\end{table*}

\begin{table*}[p] % Place the table on a separate page
    \centering
    \scriptsize
    \label{tab: example-3}
    \begin{tabular}{|p{1cm}|p{14cm}|}
        \hline
        \textbf{Model} & \textbf{Response} \\
        \hline
        Claude-2 & \begin{enumerate}
            \item Conduct qualitative research through interviews or focus groups to understand the specific barriers and facilitators to COVID-19 vaccination uptake in different socioeconomic groups within a country. This could provide insights to inform targeted interventions.
            \vspace{-1ex}
            \item Examine the impact of COVID-19 vaccination policies and public health messaging on vaccination uptake across socioeconomic groups. For example, study how vaccination requirements or incentives affect different populations.
            \item Investigate strategies to improve global collaboration and coordination in vaccine distribution and access. Research equitable pricing models, technology transfers, and policies to balance local needs with global supply. 
            \item Develop mathematical models to forecast COVID-19 vaccination coverage over time under different policy scenarios.
            \item Conduct intervention studies to test the effectiveness of community-based participatory approaches in increasing COVID-19 vaccine acceptance among disadvantaged populations.
        \end{enumerate} \\
        \hline
        Gemini-1.0 & \begin{enumerate}
            \item Investigate the impact of socioeconomic factors on vaccine hesitancy and refusal.This could be done through qualitative studies to understand the underlying reasons for vaccine hesitancy and refusal, and quantitative studies to examine the association between socioeconomic factors and vaccine uptake.
            \item Examine the role of social networks and community engagement in promoting vaccination.This could involve studying the influence of social networks on vaccination decisions, and exploring the effectiveness of community-based interventions in increasing vaccination rates.
            \item Assess the effectiveness of different strategies for addressing vaccine inequality.This could include evaluating the impact of financial incentives, targeted outreach programs, and mobile vaccination clinics on vaccination rates among underserved populations.
            \item Develop and evaluate interventions to improve the equitable distribution of vaccines.** This could involve working with governments and international organizations to develop policies and programs that ensure that vaccines are distributed fairly and equitably to all populations.
            \item Conduct longitudinal studies to track the long-term impact of socioeconomic factors on vaccination rates.This could help to identify trends and patterns in vaccination uptake over time, and inform the development of policies and interventions to address persistent disparities.
        \end{enumerate} \\
        \hline
        GPT-4 & \begin{enumerate}
            \item Impact of Vaccine Acceptance and Hesitancy on Socioeconomic Disparities: Investigate how vaccine acceptance and hesitancy, influenced by cultural, historical, and information dissemination factors, contribute to the observed socioeconomic disparities in vaccination rates. This research could further dissect how misinformation and trust in healthcare systems intersect with socioeconomic status to affect vaccine uptake.
            \item Longitudinal Studies on the Evolution of Vaccine Inequality: Conduct longitudinal studies to track changes in vaccine inequality over time, particularly as new variants emerge and booster vaccinations become more common. This research would provide insights into how socioeconomic disparities in vaccination rates evolve with the pandemic's progression and the introduction of new vaccine technologies or strategies.
            \item Role of International and Public-Private Partnerships in Addressing Vaccine Inequity: Explore the effectiveness of international collaborations and public-private partnerships in improving vaccine distribution equity. This study could examine case studies of successful collaborations and identify best practices for leveraging these partnerships to reduce disparities in vaccine access.
            \item Comparative Analysis of Vaccination Strategies and Socioeconomic Impact: Compare the vaccination strategies of different countries or regions with similar socioeconomic profiles to identify what strategies are most effective in mitigating inequalities in vaccination rates. This research could include the role of mobile vaccination units, community engagement initiatives, and targeted communication strategies in improving vaccine uptake among disadvantaged populations.
            \item Impact of Socioeconomic Factors on Post-Vaccination Outcomes: Investigate how socioeconomic factors influence outcomes after vaccination, such as breakthrough infection rates, severity of breakthrough cases, and long-term immunity. This research could help identify whether socioeconomic disparities persist even after vaccination and how these disparities can be addressed through booster doses, tailored public health strategies, and improved healthcare access.
        \end{enumerate} \\
        \hline
        GPT-3.5 & \begin{enumerate}
            \item Investigating the Impact of Socioeconomic Factors on Access and Uptake of COVID-19 Booster Vaccines: A study could focus on analyzing how socioeconomic indicators, such as GDP, human development index, and corruption perception index, influence access to and acceptance of booster doses of COVID-19 vaccines. 
            \item Comparative Analysis of Inequality in COVID-19 Vaccination Across Continents: Future research could compare and contrast the levels of inequality in COVID-19 vaccination across different continents, especially focusing on the factors contributing to variations in vaccination rates and disparities in vaccine distribution. 
            \item Longitudinal Study on the Impact of Socioeconomic Factors on COVID-19 Vaccination: A longitudinal study could be conducted to track changes in vaccination rates over time and assess how socioeconomic factors impact the progress of COVID-19 vaccination campaigns. 
            \item Exploring Public Perception and Vaccine Hesitancy Among Different Socioeconomic Groups: Future research could delve into public perceptions and attitudes towards COVID-19 vaccination among diverse socioeconomic groups, investigating factors such as education level, income, and access to healthcare.
            \item Evaluation of Intervention Strategies to Reduce Inequality in COVID-19 Vaccination: Researchers could conduct a study to evaluate the effectiveness of various intervention strategies aimed at reducing inequality in COVID-19 vaccination. By implementing targeted initiatives and measuring their impact on vaccine distribution and uptake, this research could provide evidence-based recommendations for policy interventions to promote equitable access to vaccines.
        \end{enumerate} \\
        \hline
    \end{tabular}
    \caption{Example of generated response for economics domain. Paper:
    \url{https://www.mdpi.com/2076-393X/11/7/1245}}
    \label{Table: output-3}
\end{table*}

\begin{table*}[p] % Place the table on a separate page
    \centering
    \scriptsize
    \label{tab: example-4}
    \begin{tabular}{|p{1cm}|p{14cm}|}
        \hline
        \textbf{Model} & \textbf{Response} \\
        \hline
        Claude-2 & \begin{enumerate}
            \item Further optimize the minimal active H2 relaxin analog B9-31 through additional amino acid substitutions and modifications to improve binding affinity and activity.
            \item Explore alternative methods to induce helicity and stabilize the minimal H2 relaxin analogs, such as cyclization or incorporation of other helix-inducing amino acids.
            \item Develop longer-acting analogs by modifying B9-31 with polyethylene glycol, albumin binding moieties, or other half-life extending strategies. 
            \item Test the in vivo efficacy of optimized minimal analogs in animal models of fibrosis, heart failure, and other conditions to assess therapeutic potential.
            \item Design biased analogs to selectively activate certain signaling pathways downstream of RXFP1 to optimize beneficial effects and reduce side effects.
        \end{enumerate} \\
        \hline
        Gemini-1.0 & \begin{enumerate}
            \item Investigate the potential benefits of using ECMO in combination with other therapies for severe respiratory failure. For example, ECMO could be used in combination with lung transplantation, mechanical ventilation, or extracorporeal membrane oxygenation (ECMO). This could help to improve patient outcomes by providing more comprehensive support for the lungs and other organs.
            \item Develop new strategies for managing complications associated with ECMO. Common complications of ECMO include bleeding, infection, and thrombosis. New strategies for managing these complications could help to improve patient safety and outcomes.
            \item Study the long-term effects of ECMO on patients.There is limited data on the long-term effects of ECMO on patients. Studies are needed to investigate the potential risks and benefits of ECMO in the long term.
            \item Develop new technologies to improve the performance of ECMO.New technologies could help to make ECMO more effective, safer, and easier to use. For example, new oxygenators could be developed that are more efficient and less likely to cause complications.
            \item Conduct research to better understand the mechanisms of respiratory failure.This could help to identify new targets for therapy and improve the overall management of respiratory failure.
        \end{enumerate} \\
        \hline
        GPT-4 & \begin{enumerate}
            \item Comparative Effectiveness of ECMO Configurations: Conduct randomized controlled trials comparing different ECMO cannulation strategies (e.g., dual-lumen cannula via the right internal jugular vein vs. traditional multi-cannula approaches) to assess their impact on patient outcomes, including survival rates, complication rates, and the ability to mobilize and rehabilitate during ECMO support. This research could also explore the use of image guidance for cannula placement and its impact on patient outcomes.
            \item ECMO in the Management of COVID-19-Induced ARDS: Given the limited randomized controlled trial data on the efficacy of ECMO for COVID-19-associated respiratory failure, further research is needed to define the role of ECMO in this context. Future studies could focus on identifying which patients benefit most from ECMO, the timing of initiation, and how ECMO is best integrated with other supportive therapies (e.g., antivirals, anti-inflammatory agents).
            \item Long-term Outcomes and Quality of Life Post-ECMO: Investigate the long-term outcomes of patients who have undergone ECMO support for severe respiratory failure, including physical, cognitive, and psychological recovery. This research could also assess the quality-adjusted life years (QALYs) gained through ECMO support and identify factors that predict better long-term outcomes.
            \item Optimization of Anticoagulation Strategies: Develop and test new anticoagulation protocols or agents that minimize the risks of bleeding and thrombotic complications in ECMO patients. This could include comparative studies of direct thrombin inhibitors versus unfractionated heparin, as well as the exploration of novel anticoagulants and biocompatible circuit coatings that may reduce the need for systemic anticoagulation.
            \item Integration of ECMO with Rehabilitation: Explore the feasibility, safety, and benefits of early rehabilitation protocols in ECMO-supported patients, including those on "awake ECMO." This research could assess the impact of rehabilitation on outcomes such as muscle strength, functional status at discharge, duration of hospital stay, and overall mortality. Studies could also investigate the best practices for physical therapy, nutritional support, and psychological support during ECMO.
        \end{enumerate} \\
        \hline
        GPT-3.5 & \begin{enumerate}
            \item Investigating the long-term outcomes and quality of life of patients who have undergone VV-ECMO support for respiratory failure, including factors influencing long-term recovery and rehabilitation post-ECMO.
            \item Exploring new technologies and techniques to improve the efficiency and safety of VV-ECMO, such as developing novel anticoagulation strategies or improving cannulation methods to reduce complications and enhance patient outcomes.
            \item Conducting comparative studies to evaluate the effectiveness of different ventilation strategies in conjunction with VV-ECMO support, aiming to optimize patient management and potentially improve survival rates in severe respiratory failure cases.
            \item Examining the impact of ECMO utilization during disease outbreaks, such as COVID-19, on healthcare resource allocation, patient outcomes, and healthcare system preparedness, to inform future pandemic response strategies.
            \item Investigating the role of personalized medicine in VV-ECMO patient selection and management, including the potential use of biomarkers, genetic profiling, or other advanced techniques to tailor ECMO therapy to individual patient characteristics and optimize outcomes.
        \end{enumerate} \\
        \hline
    \end{tabular}
    \caption{Example of generated response for medical domain. Paper:
    \url{https://link.springer.com/article/10.1007/s41030-023-00214-2}}
    \label{Table: output-4}
\end{table*}
\begin{table*}[p] % Place the table on a separate page
    \centering
    \scriptsize
    \label{tab: example-5}
    \begin{tabular}{|p{1cm}|p{14cm}|}
        \hline
        \textbf{Model} & \textbf{Response} \\
        \hline
        Claude-2 & \begin{enumerate}
            \item Investigate the antioxidant and oil-stabilizing effects of specific peptide sequences identified in the pepsin and trypsin hydrolysates. The paper showed the pepsin hydrolysate had higher activity but did not identify the key active peptides. Bioinformatics could be used to predict promising peptides.
            \item Test the soy protein hydrolysates in stabilizing other vegetable oils like canola, sunflower, etc. during deep frying. The effects may differ for oils with varying fatty acid compositions. 
            \item Evaluate the sensory qualities of foods fried in palm olein stabilized with the soy protein hydrolysates compared to unstabilized oil. This could include assessing odor, taste, and textural changes.
            \item Study the effects of the hydrolysates on the nutritional quality of the palm olein during repeated deep frying cycles. Parameters could i+nclude vitamin E content, formation of trans fats, and retention of essential fatty acids. 
            \item Investigate combining the soy protein hydrolysates with other antioxidants like rosemary extract or green tea to evaluate synergistic effects on stabilizing frying oils. Mixtures may provide greater protection than individual additives.
        \end{enumerate} \\
        \hline
        Gemini-1.0 & \begin{enumerate}
            \item Investigate the effect of different hydrolysis conditions on the antioxidant activity of soy protein hydrolysates.This could involve varying the enzyme concentration, temperature, pH, and hydrolysis time to determine the optimal conditions for producing hydrolysates with the highest antioxidant activity.
            \item Evaluate the antioxidant activity of soy protein hydrolysates in different food systems. This could involve adding hydrolysates to edible oils, fried foods, or other food products and measuring the oxidative stability of the products over time. This would help to determine the potential applications of soy protein hydrolysates as natural antioxidants in food products.
            \item Study the mechanism by which soy protein hydrolysates exert their antioxidant activity.** This could involve using in vitro and in vivo assays to investigate the ability of hydrolysates to scavenge free radicals, chelate metal ions, or inhibit lipid peroxidation. This information would help to elucidate the molecular mechanisms underlying the antioxidant activity of soy protein hydrolysates.
            \item Explore the potential health benefits of soy protein hydrolysates. This could involve conducting animal or human studies to investigate the effects of hydrolysates on oxidative stress, inflammation, and other health markers. This would help to determine the potential role of soy protein hydrolysates in promoting human health.
            \item Develop novel food products containing soy protein hydrolysates. This could involve incorporating hydrolysates into functional foods, such as beverages, snacks, or supplements, to improve their nutritional value and antioxidant content. This would help to increase the consumption of soy protein hydrolysates and promote their potential health benefits.
        \end{enumerate} \\
        \hline
        GPT-4 & \begin{enumerate}
            \item Comparative Analysis of Different Plant-Based Proteins on Oil Stability: Extending the research to other plant-based proteins, such as pea, hemp, and rice proteins, to compare their effectiveness in improving oil oxidative stability during frying cycles. This could provide a broader understanding of the potential of plant-based proteins in enhancing the nutritional and safety profiles of fried foods.
            \item Mechanism of Action at Molecular Level: Delving deeper into the molecular mechanisms behind the antioxidant activities of different hydrolysates. This could involve investigating the interaction between specific peptide sequences and oxidative compounds within the oil, using computational modeling and in vitro experiments to elucidate the pathways for their antioxidant actions.
            \item Impact of Hydrolysate Concentration and Frying Duration: Exploring the effect of varying concentrations of the most effective hydrolysates (e.g., from pepsin digestion) over extended frying cycles beyond 12, to determine the optimal concentration for maximum oxidative stability without affecting the sensory qualities of the oil and the fried product.
            \item Application in Industrial Frying Processes: Conducting pilot-scale studies in industrial settings to assess the feasibility and economic viability of incorporating hydrolysates into commercial frying oils. This research could include the development of guidelines for the use of hydrolysates in different types of frying operations and their impact on the quality and shelf-life of fried foods.
            \item Development of Natural Antioxidant Blends: Formulating blends of hydrolysates with other natural antioxidants (e.g., extracts from spices, herbs, or other food by-products rich in phenolic compounds) to synergistically enhance the oxidative stability of frying oils. This research could also investigate the sensory impact of these blends on fried products and consumer acceptance.
        \end{enumerate} \\
        \hline
        GPT-3.5 & \begin{enumerate}
            \item Investigating the impact of different proteases on the generation of antioxidant peptides from soy protein hydrolysates and their effectiveness in stabilizing oils during frying cycles.
            \item Exploring the structural features of peptides derived from soy protein hydrolysates to determine the key amino acids responsible for their antioxidant properties.
            \item Comparing the oxidative stability of different types of oils (e.g., sesame oil, canola oil, corn oil) when stabilized with soy protein hydrolysates under varied frying conditions.
            \item Examining the influence of hydrolyzed protein residues on the formation of secondary oxidation products in oils during frying cycles and their impact on food quality and safety.
            \item Utilizing bioinformatics tools to predict and select specific peptide sequences from soy protein hydrolysates that exhibit the highest antioxidant capacity and stability-enhancing properties in fried oils..
        \end{enumerate} \\
        \hline
    \end{tabular}
    \caption{Example of generated response for physics domain. Paper:
    \url{https://www.ncbi.nlm.nih.gov/pmc/articles/PMC10528889/pdf/foods-12-03498.pdf}}
    \label{Table: output-5}
\end{table*}

\begin{table*}[!h] % Place the table on a separate page
    \centering
    \scriptsize
    \begin{tabular}{|p{14cm}|}
        \hline
1. Test-time Adaptation of Discriminative Models via Diffusion Generative Feedback  Adapts pre-trained discriminative models to each unlabelled example in the test set using generative feedback from a diffusion model. \\
2. Adaptive Discriminative Generative Model for Object Tracking  Formulates a novel discriminative generative framework that generalizes the conventional Fisher Linear Discriminant algorithm with a generative model and renders a proper probabilistic interpretation. \\
3. Classification with Hybrid Generative/Discriminative Models  Describes a hybrid model in which a high-dimensional subset of the parameters are trained to maximize generative likelihood, and another, small, subset of parameters are discriminatively trained to maximize conditional likelihood. \\
4. Discriminative Level Set for Contour Tracking  Integrates discriminative methods into a level set framework when constructing the level set energy function. \\
5. ManiFPT  Defining and Analyzing Fingerprints of Generative Models  Formalizes the definition of artifact and fingerprint in generative models, proposes an algorithm for computing them in practice, and finally study its effectiveness in distinguishing a large array of different generative models. \\
6. Generative Models for 3D Point Clouds  Experiments with transformer encoders, latent-space flow models, and autoregressive decoders to improve the performance of point cloud latent-space generative models. \\
7. Models and Modeling    \\
8. Do text-free diffusion models learn discriminative visual representations?  Explores the possibility of a unified representation learner, a diffusion model, which addresses both generative and discriminative tasks simultaneously. \\
9. Fine-Tuning Generative Models as an Inference Method for Robotic Tasks  Investigates how to quickly adapt the sample generation of neural network models to observations in robotic tasks. \\
10. Discriminative locally document embedding  Learning a smooth affine map by approximation of the probabilistic generative structure of subspace    \\
11. Working with Deep Generative Models and Tabular Data Imputation  Provides a fair comparison of proposed methods for imputing missing values in tabular data using deep generative models. \\
12. Robust Discriminative Principal Component Analysis   \\ 
13. Generative Second Language Acquisition   \\
14. Nonlinear Models   \\
15. Understanding how Differentially Private Generative Models Spend their Privacy Budget  Analyzes how DP generative models distribute privacy budgets across rows and columns of tabular data. \\
16. Online multiple object tracking by hierarchical association of detection responses  Presents a framework for multi-pedestrian tracking using a hierarchical association of detection responses, learning both discriminative and generative appearance models online. \\
17. Two-Stage Generative Learning Objects    \\
18. Generative design games activity   \\
19. First vs second quantization  \\
20. Non-discrimination Criteria for Generative Language Models  Studies how to uncover and quantify the presence of gender biases in generative language models, deriving generative AI analogues of three well-known non-discrimination criteria from classification. \\
\hline
    \end{tabular}
    \caption{Example of background knowledge of
    \url{https://ieeexplore.ieee.org/document/10191295}}
    
    \label{tab: background}
\end{table*}

\section{Output Examples}
Our LLM generated future research output can be found in Table \ref{Table: output-1}, Table \ref{Table: output-2}, Table \ref{Table: output-3}, Table \ref{Table: output-4} and Table \ref{Table: output-5}.

\section{Model Training Cutoff}

Regarding model training cutoffs, we used GPT-3.5-turbo-0125 for GPT-3.5 (cutoff: December 2023), gpt-4-0125-preview for GPT-4 (cutoff: December 2023), anthropic.claude-v2 for Claude 2 (cutoff: August 2023), and gemini-1.0-pro-002 (cutoff: February 2023), Llama 3.3 70B Instruct (cutoff: December 2023), gpt-4o-2024-08-06 for GPT4o (cutoff: October 2023), claude-3-opus-20240229 for Claude 3 (cutoff: August 2023), gemini-2.5-pro-preview-05-06 for Gemini-2.5 (cutoff: December 2025).

\section{Evaluation on Newly Released Models}
\label{Sec: new_model}
To evaluate the performance of recently released language models, we constructed a supplementary Computer Science dataset comprising 200 papers published between January and March 2025. These papers were sourced from \textsc{S2ORC} using the same English-language, PDF-availability, and \emph{Future Work} section filters as described in Section~\ref{Sec: dataset}. Annotation was carried out following the same protocol as for the main dataset (Appendix~\ref{Appendix: A}), ensuring consistency in labeling. This new dataset forms the basis for the IAScore results reported for GPT-4o, Claude-3, Gemini-2.5, and Llama-70B (Figure~\ref{fig: iascore_new}). We plan to automate the process to automatically evaluate the IAScore for newly released models.

\begin{table*}[ht]
\centering
\begin{tabular}{lcc}
\toprule
\textbf{Domain} & \textbf{Best Performing LLM} & \textbf{Worst Performing LLM} \\
\midrule
Chemistry        & Claude-2       & Gemini-1.0 \\
Economics        & Claude-2       & Gemini-1.0 \\
Computer Science & GPT-4 (IAScore), Claude-2 (Distinctness) & Gemini-1.0 \\
Medical          & GPT-4          & Gemini-1.0 \\
Physics          & GPT-4 (IAScore), Claude-2 (Distinctness) & Gemini-1.0 \\
\bottomrule
\end{tabular}
\caption{Summary of best and worst performing LLMs across domains based on alignment (IAScore) and diversity (Distinctness).}
\label{tab:llm_performance}
\end{table*}

\begin{table*}[ht]
\centering
\begin{tabular}{lccc}
\toprule
\textbf{Model} & \textbf{Novelty} & \textbf{Relevance} & \textbf{Feasibility} \\
\midrule
Claude-2              & Moderate      & Moderate         & High \\
GPT-4                 & High          & Very High        & Very High \\
Claude-2 (+BG)        & Higher        & Slightly Lower   & Slightly Lower \\
GPT-4 (+BG)           & Highest       & Highest          & Highest \\
\bottomrule
\end{tabular}
\caption{Human evaluation results in Computer Science domain. (+BG) indicates use of background knowledge.}
\label{tab:human_eval_summary}
\end{table*}


% \appendix

% \section{Example Appendix}
% \label{sec:appendix}

% This is an appendix.

\end{document}